\documentclass[letterpaper]{article} 
\usepackage{aaai25}  
\usepackage{times}  
\usepackage{helvet}  
\usepackage{courier}  
\usepackage[hyphens]{url}  
\usepackage{graphicx} 
\usepackage{natbib}  
\usepackage{caption} 
\usepackage{algorithm}
\usepackage{algorithmic}
\usepackage{amsmath}
\usepackage{booktabs}
\usepackage{amssymb}
\usepackage{enumitem}
\usepackage{etoolbox}

\usepackage{newfloat}
\usepackage{listings}
\DeclareCaptionStyle{ruled}{labelfont=normalfont,labelsep=colon,strut=off} 
\floatstyle{ruled}
\newfloat{listing}{tb}{lst}{}
\floatname{listing}{Listing}
\nocopyright 

\title{Uncovering Bias Paths with LLM-guided Causal Discovery: An Active Learning and Dynamic Scoring Approach}

\author {
    Khadija Zanna\textsuperscript{\rm 1},
    Akane Sano\textsuperscript{\rm 1}
}
\affiliations {
    \textsuperscript{\rm 1}Department of Electrical and Computer Engineering, Rice University\\
    khzanna@rice.edu, Akane.Sano@rice.edu
}

\usepackage{bibentry}

\begin{document}

\maketitle

\begin{abstract}
Ensuring fairness in machine learning requires understanding how sensitive attributes like race or gender causally influence outcomes. Existing causal discovery (CD) methods often struggle to recover fairness-relevant pathways in the presence of noise, confounding, or data corruption. Large language models (LLMs) offer a complementary signal by leveraging semantic priors from variable metadata. We propose a hybrid LLM-guided CD framework that extends a breadth-first search strategy with active learning and dynamic scoring. Variable pairs are prioritized for querying using a composite score combining mutual information, partial correlation, and LLM confidence, enabling more efficient and robust structure discovery. To evaluate fairness sensitivity, we introduce a semi-synthetic benchmark based on the UCI Adult dataset, embedding domain-informed bias pathways alongside noise and latent confounders. We assess how well CD methods recover both global graph structure and fairness-critical paths (e.g., sex→education→income). Our results demonstrate that LLM-guided methods, including our active, dynamically scored variant, outperform baselines in recovering fairness-relevant structure under noisy conditions. We analyze when LLM-driven insights complement statistical dependencies and discuss implications for fairness auditing in high-stakes domains.

\end{abstract}

%

\section{Introduction} \label{sec:introduction}

Bias in machine learning (ML) systems affects decisions in hiring, lending, education, and healthcare. These biases often emerge through indirect, structural pathways where sensitive attributes (e.g., race or gender) influence outcomes via proxies or confounded relationships \cite{graetz2022structural}. Standard fairness audits rely on statistical disparity metrics, but such metrics overlook how bias propagates causally \cite{chinta2025ai}. Without structural insight, interventions may be ineffective or misleading.

Causal discovery (CD) offers tools to identify such pathways, distinguishing genuine effects from those introduced by confounders or measurement artifacts. While recent methods apply structural causal models or fairness constraints to audit ML systems, classical CD techniques often fail under noisy data, latent confounding, or incomplete metadata. These limitations are acute in fairness contexts where both precision and interpretability are critical \cite{takayama2024integrating}.

Large language models (LLMs) have emerged as promising tools for CD, using their vast semantic knowledge to infer causal directions between variables from textual metadata \cite{le2024multi,vashishtha2023causal}. Prior work has used LLMs to infer causal links or orderings, and to reduce query cost via strategies like breadth-first search (BFS) \cite{jiralerspong2024efficient}. However, naive LLM use risks over or under attributing causality, especially involving sensitive attributes, raising concerns about spurious fairness conclusions.

Building on prior work, we introduce a fairness-driven CD framework that enhances the BFS approach with active learning (AL) and dynamic scoring. Our method prioritizes variable pairs using mutual information (MI), partial correlation (PCorr), and LLM-derived confidence, while discounting redundant queries and focusing on informative regions via a history-based weight. Unlike exhaustive querying or fixed-order search, our adaptive strategy balances semantic and statistical cues to recover fairness-relevant paths more efficiently.

A key challenge in evaluating CD methods in fairness contexts is the absence of real-world datasets with known ground-truth causal structures involving sensitive attributes \cite{loftus2018causal}. Most, like the UCI Adult dataset, offer rich features but lack verified causal graphs. To enable robust evaluation, we construct a semi-synthetic benchmark from the UCI Adult dataset, embedding a known fairness-relevant causal graph with noise, corruption, and confounding. We assess how well CD methods recover both global structure and key fairness pathways (e.g., \textit{sex} $\rightarrow$ \textit{education} $\rightarrow$ \textit{income}).

We also conduct a hyperparameter sensitivity analysis to examine how LLM-specific factors such as temperature, query prioritization weights, and stopping criteria affect CD performance. These findings offer practical insights into how LLM behavior and prompting influence graph reconstruction accuracy and fairness-path recovery. Our approach is designed for robustness, interpretability, and social accountability, enabling stakeholders to trace how outcomes are structurally linked to sensitive variables and to intervene in meaningful ways.

Our key contributions:
\begin{itemize}
    \item A hybrid LLM-CD framework with active learning and dynamic scoring to enhance fairness-aware structure discovery.
    \item A benchmark for fairness-sensitive CD grounded in real-world semantics and controlled ground truth.
    \item An evaluation of pathway recovery across methods, highlighting conditions where LLM-guided discovery improves fairness diagnostics.
    \item A sensitivity analysis on LLM parameters and their influence on recovery quality.

\end{itemize}

\section{Related Work}

\textbf{Statistical and Optimization-Based CD.}  
CD methods span statistical, optimization-based, and more recently, LLM-guided approaches \cite{long2025survey}. Classical techniques such as PC \cite{spirtes1991algorithm} and GES \cite{meek1997graphical} rely on conditional independence tests and strong assumptions (e.g., faithfulness), often failing under latent confounding or limited data. Optimization-based methods like NOTEARS \cite{zheng2018dags} and DAGMA \cite{bello2022dagma} offer greater precision but are computationally expensive and sensitive to hyperparameters, making them less practical for fairness-critical contexts.

\textbf{LLM-Guided CD.}  
Recent LLM-based approaches use variable semantics to infer structure, with Kıcıman et al. \cite{kiciman2023causal} proposing pairwise causal queries from metadata. However, such methods scale poorly due to quadratic query complexity. Others combine LLM-inferred causal orders with classical CD \cite{vashishtha2023causal}, or use multi-agent prompting strategies \cite{khatibi2024alcm, le2024multi}. Kampani et al. \cite{kampani2024llm} generate LLM-based priors refined through NOTEARS.  
Takayama et al. \cite{takayama2024integrating} introduced statistical causal prompting, where LLMs enhance statistical methods by providing priors or plausible causal orders. Jiralerspong et al. \cite{jiralerspong2024efficient} proposed a BFS strategy to reduce query complexity, but rely on fixed pair ordering, making them susceptible to early decision errors.
  
We build on this line by introducing AL and a dynamic scoring mechanism to adaptively prioritize informative variable pairs using MI, PCorr, and LLM confidence. Our method dynamically balances semantic and statistical signals, in contrast to prior work that uses LLMs for one-time priors or uniform querying. We focus on observational CD methods to maintain comparability with LLM-based approaches, leaving integration with interventional or experimental methods (e.g., do-calculus, IV-based CD) for future work.

\textbf{Fairness and Causal Pathways.}  
Beyond structural accuracy, recent work has emphasized the importance of CD for fairness auditing. Causal fairness frameworks \cite{nabi2018fair, kilbertus2017avoiding, zhang2018fairness, chiappa2019path} focus on identifying discriminatory pathways, e.g., \textit{sex} $\rightarrow$ \textit{education} $\rightarrow$ \textit{income} that explain how bias propagates through mediation or proxy variables. Our method is explicitly tuned to recover such fairness-relevant paths with greater robustness under noise and limited samples, enabling more actionable fairness assessments.

\textbf{Social and Regulatory Context.}  
Beyond computer science, our work connects to social science perspectives. Structural discrimination and intersectionality highlight how compounding disadvantage flows through proxy variables, not just direct links \cite{crenshaw2013demarginalizing, bonilla1997rethinking}. Regulatory frameworks such as GDPR’s “right to explanation” and EEOC compliance requirements also motivate interpretable, path-level audits \cite{kaminski2021right}.

\textbf{Causal Inference and Accountability.}  
We also draw from causal inference frameworks like the potential outcomes model \cite{rubin1974estimating, imbens2015causal} and matching-based fairness analysis \cite{stuart2010matching}, which stress structural validity and transparency of assumptions. Our hybrid design reflects these values by surfacing interpretable causal hypotheses grounded in both observed data and language-based priors. Finally, we respond to calls for auditable AI systems through transparency-centered work on model documentation and sociotechnical accountability \cite{mitchell2019model, selbst2019fairness}.


\section{Proposed LLM-Based Method} \label{sec:Method}

We build on the BFS-based CD framework of \cite{jiralerspong2024efficient}, where an LLM is queried iteratively to construct a causal graph. The method begins by identifying independent root variables and expands outward via LLM-guided queries, treating each confirmed edge as a BFS traversal step. To ensure acyclicity, edges are added only if they do not create cycles. While more efficient than exhaustive querying, this approach treats all variable pairs uniformly and can waste queries on uninformative or redundant relationships.

\subsection{Active Learning and Dynamic Scoring for Efficient Querying}

\begin{figure*}[t]
  \centering
  \includegraphics[scale=0.32]{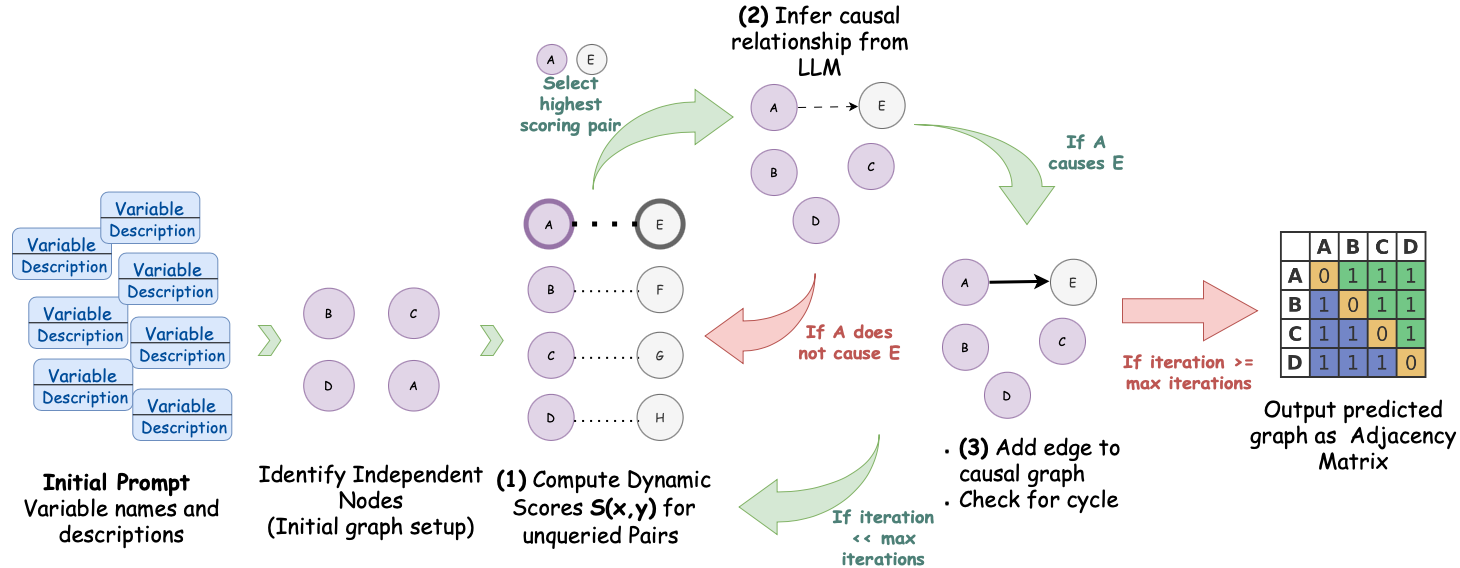}
  \caption{Overview of the proposed LLM-guided BFS framework with dynamic scoring and AL.}
  \label{fig:method_figure}
\end{figure*}

We introduce an AL strategy that selects variable pairs based on a dynamic informativeness score. This score integrates statistical signals, model confidence, and query history, prioritizing pairs likely to yield informative causal judgments. The query loop continues until a maximum iteration or informativeness threshold is reached. Figure~\ref{fig:method_figure} summarizes the overall architecture.

\subsubsection{Dynamic Scoring Mechanism}

Each unqueried pair $(x, y)$ receives a composite score:

\begin{equation}
\label{eq:dynamic_score}
\begin{split}
S(x, y) =\ & w_{\text{stat}} \cdot \text{StatScore}(x, y) + \\
          & w_{\text{conf}} \cdot \text{LLMConf}(x, y) + 
            w_{\text{hist}} \cdot \text{HistScore}(x, y)
\end{split}
\end{equation}

Each term incorporates a distinct signal:
\begin{itemize}
    \item \textbf{Statistical Score:} Combines MI and PCorr, measuring both linear and non-linear dependencies:
    \begin{equation}
        \text{StatScore}(x, y) = \frac{\text{MI}(x, y) + \text{PCorr}(x, y)}{2}
    \end{equation}

PCorr is computed conditionally on the current discovered parent sets, following the iterative conditioning procedure of the PC algorithm. 
    
    \item \textbf{LLM Confidence Score:} Reflects the certainty of the model's previous response for a given pair, where higher token-level confidence leads to higher scores:
    \begin{equation}
        \text{LLMConf}(x, y) = \frac{1}{1 + e^{-\text{confidence}}}
    \end{equation}

    \item \textbf{Query History Score:} Penalizes repeated queries to encourage broader exploration:
    \begin{equation}
        \text{HistScore}(x, y) = \frac{1.5}{1 + \text{query\_count}(x, y)}
    \end{equation}
\end{itemize}

The weights $w_{\text{stat}}, w_{\text{conf}}, w_{\text{hist}}$ are treated as tunable hyperparameters. These weights are optimized via Bayesian optimization (see Section~\ref{sec:experiments}), allowing the framework to adapt scoring behavior to different graph structures or domain characteristics.

\subsubsection{Querying the LLM}

At each step, the pair $(x^*, y^*)$ with the highest score is selected:

\begin{equation}
(x^*, y^*) = \arg\max_{(x, y) \in \text{Unqueried}} S(x, y)
\end{equation}

The LLM is queried using a simple natural language prompt:

\begin{listing}[H]
\caption{Simplified LLM Query Format}
\label{lst:prompt}
\begin{lstlisting}
System: You are a domain expert.
User: You assist the expert in evaluating possible causal links.

Metadata: Each variable is described using a short name and a natural language description.

Query: Does VAR_A cause VAR_B?
Respond with <Answer>Yes</Answer> or <Answer>No</Answer>.
\end{lstlisting}
\end{listing}

Domain-specific instructions and variable descriptions are included to contextualize each query. See Appendix for full example prompts.

If the LLM returns ``Yes'' and adding $x^* \rightarrow y^*$ does not form a cycle, the edge is added to the graph. To estimate the LLM uncertainty, we extract token-level log-probabilities and compute the average probability of the model’s response (“Yes” or “No”). This score, scaled to [0,1], informs the dynamic scoring function. If log-probs are unavailable, a default confidence of 0.5 is used. The process terminates after a query limit or when scores fall below a threshold.

To ensure the output is a valid DAG, each edge addition is immediately followed by a cycle check. If the graph becomes cyclic, the addition is reversed.

The final causal graph is returned both as a directed edge list and an adjacency matrix:

\begin{equation}
A(i, j) = 
\begin{cases} 
1 & \text{if } X_i \rightarrow X_j \text{ is predicted}, \\
0 & \text{otherwise}.
\end{cases}
\end{equation}

The entire procedure is conducted in a multi-turn chat session, allowing the LLM to maintain contextual awareness of previously inferred relationships and rationales.

\section{Fairness Evaluation via Pathway Analysis}

We assess the fairness utility of learned causal graphs by analyzing how sensitive attributes \(S\) (e.g., \texttt{race}, \texttt{sex}) influence outcomes \(Y\) (e.g., \texttt{income}). Unlike statistical fairness metrics, pathway analysis provides a causal perspective by distinguishing direct, mediated, and spurious effects \cite{pearl2009causality, pearl2022direct}.

\paragraph{Path Classification} 
We enumerate all directed paths from \(S\) and classify them as:
\begin{itemize}[leftmargin=*]
    \item \textbf{Direct:} Edges \(S \rightarrow Y\).
    \item \textbf{Indirect:} Paths \(S \rightarrow \cdots \rightarrow Y\) via mediators.
    \item \textbf{Spurious:} Paths involving \(S\) that do not reach \(Y\).
\end{itemize}
Comparing these across methods and against the ground-truth graph reveals how well fairness-relevant mechanisms are recovered.

\paragraph{Effect Decomposition} 
We estimate the causal effects of \(S\) on \(Y\) using either structural equations or interventional estimators:
\begin{align*}
    DE &: \text{Direct effect (via } S \rightarrow Y\text{)} \\
    IE &: \text{Indirect effect (via mediators)} \\
    TE &= DE + IE
\end{align*}
We normalize by outcome variance to obtain:
\begin{equation}
    C_{\text{bias}} = \frac{TE}{\text{Var}(Y)}
\end{equation}
Here, \(C_{\text{bias}}\) quantifies the fairness-relevant contribution of \(S\) to \(Y\), enabling method comparison across datasets. Higher values indicate a greater risk of bias propagation.

\section{Datasets} \label{sec:data}

We evaluate our framework on three benchmark networks of varying size and realism.

\paragraph{Synthetic Adult-Based Network.}  
We construct a semi-synthetic dataset based on the UCI Adult dataset \cite{kohavi1996scaling}, a standard benchmark in fairness research. The graph includes 15 original variables, with causal edges informed by prior work \cite{kilbertus2017avoiding, nabi2018fair}. Demographics (e.g., \texttt{race}, \texttt{sex}) influence education and occupation, which in turn affect income, both directly and via mediators such as capital gains. Age and marital status act as confounders or mediators.

To simulate structural bias, we inject direct edges from \texttt{race} and \texttt{sex} to \texttt{income}, alongside indirect paths (e.g., \texttt{sex} $\rightarrow$ \texttt{education} $\rightarrow$ \texttt{income}). These allow evaluation of direct, indirect, and total effects. Details on graph construction and data generation are in the Appendix.

\paragraph{Child Causal Network.}  
A 20-node, 25-edge Bayesian network modeling clinical, environmental, and parental factors in congenital heart disease \cite{spiegelhalter1993bayesian}. Variables include both categorical and continuous types.

\paragraph{Neuropathic Pain Network.}  
A large-scale clinical graph with 221 nodes and 770 edges capturing pathophysiological and symptomatic relationships in neuropathic pain diagnoses, derived from real-world patient data \cite{tu2019Neuropathic}.

Since the latter two datasets lack sensitive attributes, we focus on structural accuracy rather than fairness.

\section{Experiments} \label{sec:experiments}

We evaluate our LLM-based CD framework across the benchmark datasets, focusing on graph accuracy and fairness-relevant path recovery under varying noise and hyperparameter settings.

\subsection{Baselines}

We compare against the following methods:

\begin{itemize}[leftmargin=*]
    \item \textbf{PC Algorithm} \cite{spirtes1991algorithm}: Constraint-based method using independence tests.
    \item \textbf{GES} \cite{meek1997graphical}: Score-based approach optimizing a score function (e.g., BIC) via forward-backward search.
    \item \textbf{NOTEARS} \cite{zheng2018dags}: Continuous optimization using a differentiable acyclicity constraint.
    \item \textbf{DAGMA} \cite{bello2022dagma}: Neural network-based CD with sparsity and acyclicity constraints.
    \item \textbf{LLM Pairwise} \cite{kiciman2023causal}: Uses LLM to infer pairwise causality from metadata.
    \item \textbf{LLM BFS} \cite{jiralerspong2024efficient}: Explores graph structure via LLM-guided breadth-first querying.
\end{itemize}

All baselines are run using the open-source code from \cite{jiralerspong2024efficient}.

\subsection{Experimental Setup}

We use Bayesian optimization (\texttt{gp\_minimize}) with a Gaussian Process surrogate to tune key hyperparameters \cite{mockus1994application, snoek2012practical}:
\begin{itemize}[leftmargin=*]
    \item \textbf{Scoring Weights:} Weights for MI, PCorr, and query history (Eq.~\ref{eq:dynamic_score})
    \item \textbf{Score Threshold:} Minimum score required for querying a pair
    \item \textbf{LLM Temperature:} Controls randomness in responses
    \item \textbf{Max Iterations:} AL rounds per run
\end{itemize}

Each query uses a multi-turn GPT-4 (0125-preview) prompt with prior discoveries and variable metadata (Section \ref{sec:Method}). A directed edge \(X \rightarrow Y\) is added if the reply contains \texttt{<Answer>Yes</Answer>} and does not create a cycle. If ambiguous, no edge is added. Confidence scores are derived from the average probability of top-5 tokens based on log-probs; defaulting to 0.5 if unavailable.

Trials are divided into chunks across hyperparameter space. We report results from the best-performing configuration for each run. Full prompt design and search ranges are detailed in the Appendix.

We evaluate all methods on four synthetic variants of the Adult-based dataset (Section~\ref{sec:data}), varying random seeds to preserve the underlying structure while introducing independent data realizations. Fairness analysis targets paths from \texttt{sex}, \texttt{race}, and \texttt{age} to \texttt{income}. Structural accuracy is also assessed on the Child and Neuropathic networks, with LLM-based methods averaged over five independent runs.

\subsection{Evaluation Metrics}

We assess reconstruction accuracy using standard metrics:

\begin{itemize}[leftmargin=*]
    \item \textbf{Precision \& Recall:} Defined as $\text{Precision} = \frac{\text{TP}}{\text{TP + FP}}$ and $\text{Recall} = \frac{\text{TP}}{\text{TP + FN}}$.
    \item \textbf{F1 Score:} Harmonic mean of precision and recall.
    \item \textbf{Edge Count:} Number of predicted edges vs. ground truth.
    \item \textbf{Acyclicity:} Ensures output is a valid DAG.
    \item \textbf{Normalized Hamming Distance (NHD):} $\text{NHD} = \frac{\text{Mismatches}}{n^2}$, where $n$ is the number of nodes.
    \item \textbf{Adjacency Accuracy:} $\text{Accuracy} = \frac{\text{Correct entries}}{n^2}$.
\end{itemize}

\section{Results and Analyses}

\subsection{Graph Performance Comparison}

\begin{table*}[ht]
\label{tb:performance}
\begin{center}
\begin{scriptsize}
\begin{tabular}{l||l||ccccccc}
\toprule
   \textbf{Adult-based Synthetic} &\textbf{Method} & \textbf{Acc. ($\uparrow$)} & \textbf{F1 Score ($\uparrow$)} & \textbf{Precision} & \textbf{Recall} & \textbf{NHD ($\downarrow$)}  & \textbf{NHD Ratio ($\downarrow$)} & \textbf{\# Pred. Edges} \\
   
\cmidrule{2-9}

  (15 nodes, 28 edges) & PC & 0.239 & 0.382 & 0.352 & 0.420 & 0.193 & 0.743 & 33 \\
  & GES & 0.296 & 0.473 & 0.368 & 0.580 & 0.203 & 0.782 & 44 \\
  & NOTEARS ($\lambda=0.01$) & 0.021 & 0.039 & 0.035 & 0.045 & 0.260 & 0.650 & 27 \\
  & DAGMA ($\lambda=0.05$) & 0.099 & 0.180 & 0.141 & 0.250 & 0.283 & 0.794 & 50 \\
  & LLM Pairwise & 0.307 & 0.470 & 0.331 & \textbf{0.813} & 0.253 & \textbf{0.530} & 69 \\
  & LLM BFS & 0.299 & 0.456 & 0.332 & 0.750 & 0.305 & 0.539 & 64 \\
  
\cmidrule{2-9}
  & \textbf{Proposed Method} & \textbf{0.413} & \textbf{0.585} & \textbf{0.792} & 0.464 & \textbf{0.109} & \textbf{0.415} & 17 \\

\midrule

  \textbf{Child} & PC & 0.146 & 0.255 & 0.273 & 0.239 & 0.097 & 0.745 & 22 \\
  (20 nodes, 25 edges) & GES & 0.206 & 0.341 &  0.438 & 0.279 & 0.119 & 0.659 & 16 \\
  & NOTEARS & 0.216 & 0.356 & 0.403 & 0.319 & \textbf{0.080} & 0.644 & 20 \\
  & DAGMA & 0.179 & 0.304 & 0.333 & 0.279 & 0.089 & 0.696 & 21 \\
  & LLM Pairwise & 0.130 & 0.229 & 0.144 & \textbf{0.559} & 0.235 & 0.770 & 97 \\
  & LLM BFS & 0.150 & 0.261 & 0.286 & 0.240 & 0.085 & 0.739 & 21 \\

\cmidrule{2-9}
    & \textbf{Proposed Method} & \textbf{0.364} & \textbf{0.533} & \textbf{0.601} & 0.479 & 0.082 & \textbf{0.467} & 20 \\

\midrule

  \textbf{Neuropathic} & PC & 0.041 & 0.078 & 0.092 & 0.068 & \textbf{0.025} & 0.922 & 563 \\
  (221 nodes, 770 edges) & NOTEARS & 0.022 & 0.044 & 0.500 & 0.023 & 0.334 & 0.955 & 36 \\
  & DAGMA & 0.020 & 0.039 & 0.421 & 0.021 & 0.351 & 0.960 & 38 \\
  & LLM BFS & 0.000 & 0.000 & 0.000 & 0.000 & 0.903 & 1.000 & 43 \\

\cmidrule{2-9}
  & \textbf{Proposed Method} & \textbf{0.073} & \textbf{0.136} & \textbf{0.690} & \textbf{0.075} & 0.109 & \textbf{0.864} & 84 \\

\bottomrule
\end{tabular}
\end{scriptsize}
\end{center}

\caption{Performances of Methods on the Adult-based Synthetic, Child, and Neuropathic Causal Networks. All reported values are averaged over four random seeds. Metrics exhibit minimal variance across runs ($\pm2\times10^{-4}$ in F1 score).}


\end{table*}

\paragraph{Synthetic Adult-based Network.} Our method achieves top scores across F1 (0.585), accuracy (0.413), precision (0.792), and NHD (0.109), recovering a compact graph (17 edges) with strong alignment to ground truth. LLM Pairwise attains high recall (0.813) but overpredicts, while GES and PC show moderate F1 but higher NHDs. Optimization-based methods struggle with fairness structure.

\paragraph{Child Network.} Again, our method leads in F1 (0.533), accuracy (0.364), and NHD ratio (0.467), balancing edge precision and coverage. NOTEARS achieves slightly better raw NHD but worse recall. LLM Pairwise finds many indirect effects, but at the cost of low precision.

\paragraph{Neuropathic Network.} Despite the challenge of 221 nodes, our method yields highest F1 (0.136) and precision (0.690), outperforming baselines. All baselines degrade sharply; LLM BFS fails entirely (F1 = 0). Dynamic scoring enables better query selection under sparsity.

\paragraph{Scalability.} Our method scales better than NOTEARS, DAGMA, or LLM Pairwise while offering greater structural fidelity. Complexity analysis (detailed in Appendix) shows favorable scaling theoretically in large sparse graphs, common in real-world networks. GES and LLM Pairwise were excluded from Neuropathic due to resource limits.

\paragraph{Reproducibility.} LLM-BFS and Pairwise underperform vs. \citet{jiralerspong2024efficient}, likely due to (1) GPT-4 updates and nondeterminism, (2) token limits or message handling, and (3) lack of access to original variable descriptions for the Neuropathic dataset, affecting metadata quality. Our implementation adheres to the BFS protocol and exposes its limitations under scale, motivating our enhancements.

\subsection{Fairness Pathway Recovery}

We evaluate fairness path recovery on the Adult-based synthetic graph, where the true structure contains 2 direct and 25 indirect paths from \texttt{sex}, \texttt{age}, and \texttt{race} to \texttt{income}. TE = 4.89 and \( C_{\text{bias}} = 28.46 \).

\begin{figure}[ht]
\centering
\includegraphics[width=0.9\columnwidth]
{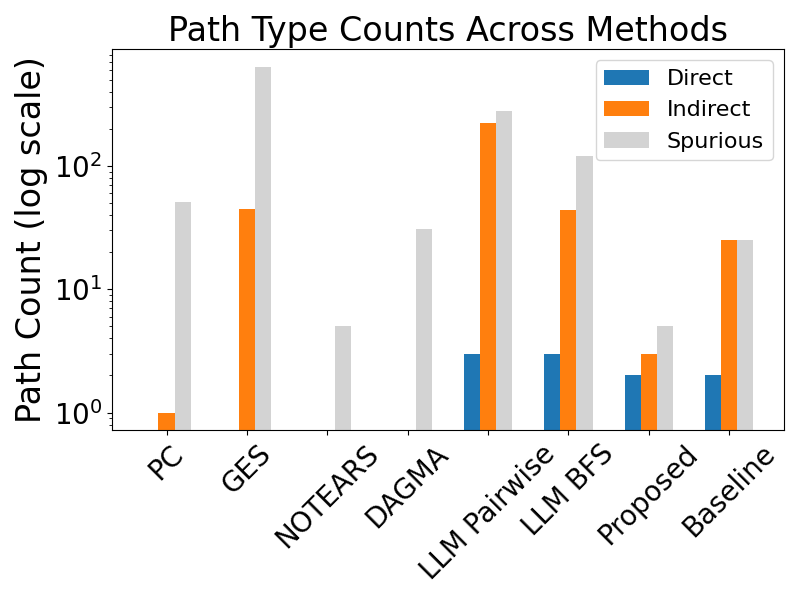} 
\caption{Log-scaled path counts (direct, indirect, spurious) per method.}
\label{fig:log_path_counts}
\end{figure}

Our method uniquely recovers both true direct paths (from \texttt{sex} and \texttt{race}) while omitting the spurious \texttt{age} path, showing strong fairness alignment. LLM methods over-attribute (3 direct paths) due to LLM priors, inflating \( C_{\text{bias}} \). PC and GES miss direct effects; NOTEARS and DAGMA capture none, yielding near-zero \( C_{\text{bias}} \).

\begin{figure}[ht]
\centering
\includegraphics[width=0.99\columnwidth]
{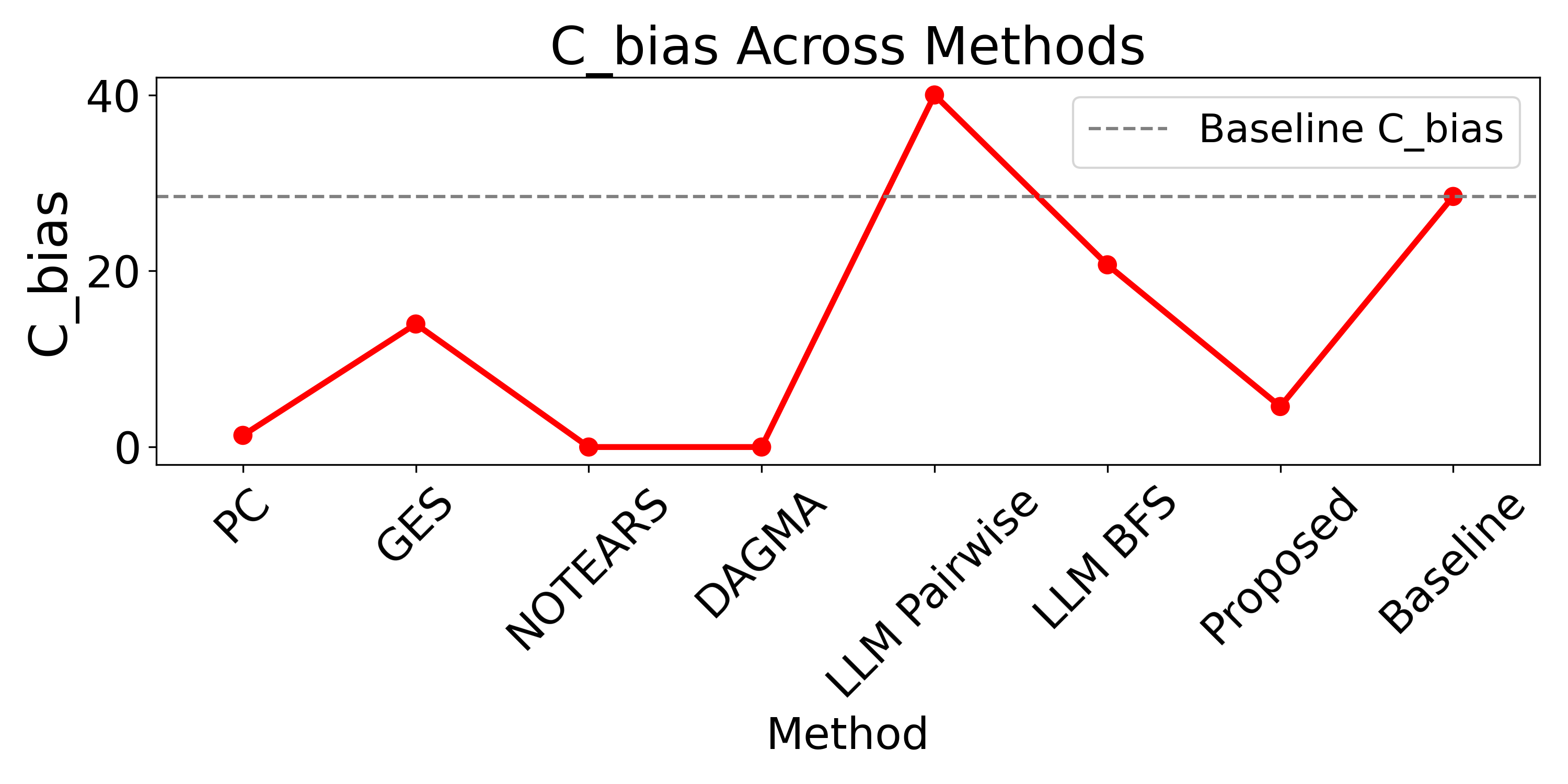} 
\caption{Normalized bias (\( C_{\text{bias}} \)) across methods.}
\label{fig:c_bias_counts}
\end{figure}

Overall, our method prioritizes high-confidence discovery, avoiding false positives and maintaining robustness across seeds as reflected by its consistently low standard deviation on key metrics such as indirect path count, total effect, and normalized bias as seen in the table in Appendix. Its conservative nature may underestimate bias but enhances reliability in fairness-critical settings.

\subsection{Ablation and Parameter Influence}

Bayesian optimization trials were analyzed via Random Forest regression to assess parameter impact on F1 score.

\begin{figure}[ht]
\centering
\includegraphics[width=\linewidth]{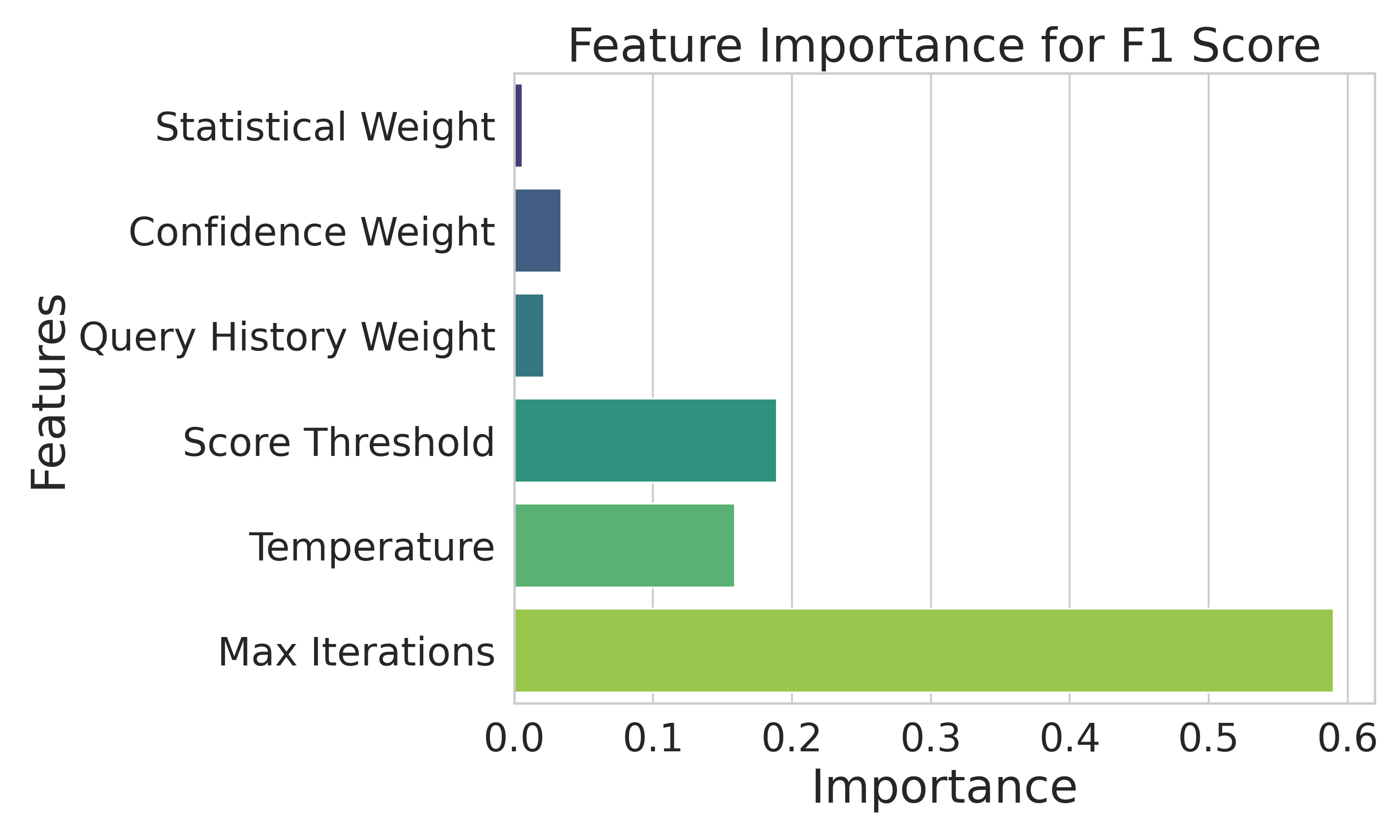}
\caption{Hyperparameter importance on F1 score.}
\label{fig:feature-importance}
\end{figure}

\begin{figure}[ht]
\centering
\includegraphics[width=\linewidth]{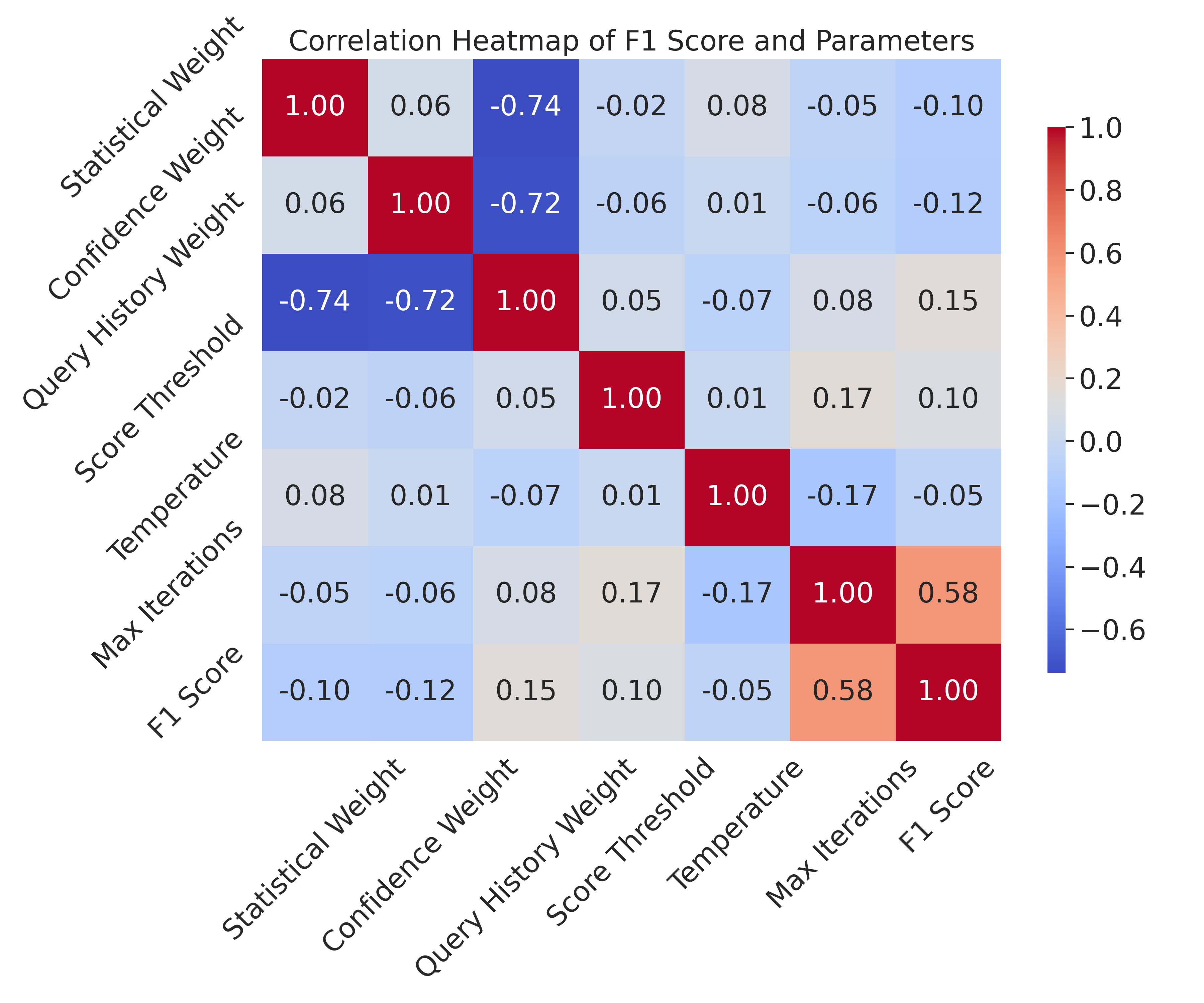}
\caption{F1 vs. hyperparameter correlation (Adult).}
\label{fig:corr-heatmap}
\end{figure}

\paragraph{Findings.} Max iterations have the strongest effect, suggesting deeper AL improves recovery. Score threshold and temperature also influence performance. Correlation analysis (Figure~\ref{fig:corr-heatmap}) shows anti-correlation among scoring weights, indicating competitive trade-offs. In small graphs, deeper querying matters more; in large ones (e.g., Neuropathic), MI/PCorr dominate (details in Appendix). This highlights how our dynamic scoring mechanism could adapt to scale, favoring semantic guidance in small graphs and statistical grounding in larger ones.

\section{Discussion}

Our findings demonstrate that combining LLM-based CD with AL and dynamic scoring improves both structural accuracy and fairness sensitivity across diverse datasets. The proposed method consistently outperforms statistical, LLM-only, and hybrid baselines on both small and large networks.

Dynamic scoring balances LLM semantic priors with empirical signals, adapting to graph complexity and data quality. In early iterations, LLM judgments guide exploration when statistical signals are weak. As more information is gathered, mutual information and partial correlation become dominant, reflecting a shift from semantic exploration to empirical refinement. AL enhances efficiency by prioritizing uncertain, low-redundancy variable pairs, avoiding the limitations of fixed-order strategies.

Our method uniquely recovers fairness-critical paths, e.g., \texttt{sex} $\rightarrow$ \texttt{education} $\rightarrow$ \texttt{income}—while avoiding spurious ones like \texttt{age} $\rightarrow$ \texttt{income}, often introduced by LLM-only baselines. This improves fairness diagnostics for impact assessments in domains like hiring or lending. While our fairness evaluation relies on a semi-synthetic benchmark, this is necessary as real-world datasets lack ground-truth causal graphs for sensitive attributes and outcomes.

However, our results also underscore broader concerns: CD methods, including ours, may introduce spurious or incomplete structures if used without scrutiny. In fairness contexts, such errors carry high stakes. False positives (e.g., unjustified bias paths) can trigger overcorrections, while false negatives obscure actual harms. We find that even underperforming methods can appear fairness-aware using naive metrics like $C_{\text{bias}}$, which may be inflated by erroneous causal paths. This highlights the need for path-level interpretability rather than summary statistics alone.

To mitigate these risks, our approach integrates LLMs as prior-informed agents within a selective AL loop. This reduces unnecessary queries, avoids early commitment errors, and improves efficiency over fixed-query strategies such as BFS. Like traditional DAG methods, our framework embraces uncertainty and treats CD as hypothesis generation. 
 
While it is argued that LLMs may not perform causal reasoning per se, we treat them as semantic priors filtered through statistical signals and query history. This is analogous to expert elicitation—subject to bias, but useful when paired with empirical validation. As argued by \cite{imbens2020potential} and \cite{mccoy2023much}, epistemic uncertainty is intrinsic to CD. LLM-guided proposals can still surface plausible structures, especially in fairness-critical settings. Prior work has emphasized this role \cite{kiciman2023causal}; we build on it with a dynamic querying process that improves recovery while reducing false positives in bias pathways.

Our broader aim is to support interpretable, efficient, and socially aware fairness diagnostics, offering complementary insight to notions like demographic parity or counterfactual fairness. By uncovering how sensitive variables influence outcomes through direct and mediated paths, our method moves beyond outcome disparities toward causal insight. This aligns with calls for path-level transparency in AI audits from fairness, accountability, and legal communities. We see this work as a step toward embedding causal reasoning into policy evaluations, model and risk assessments, tools needed to build more trustworthy socio-technical systems. 

This approach provides actionable insights for non-technical stakeholders. By surfacing fairness-relevant pathways, the method enables practitioners, compliance teams, or policy analysts to trace how sensitive attributes indirectly shape outcomes. These insights can guide interventions, from revising decision rules to informing accountability processes in hiring, lending, or regulatory oversight.

\section{Limitations}

\paragraph{Simplified Synthetic Causal Structure.}  
Our synthetic graph models \texttt{income} as a terminal node to isolate fairness pathways, aiding evaluative clarity. However, this excludes realistic downstream effects (e.g., on health, opportunity), omitting feedback or temporal dynamics. Future work should extend this to longitudinal or post-outcome settings.

\paragraph{Hyperparameter Sensitivity and Compute Costs.}  
Our method requires tuning multiple hyperparameters (e.g., score weights, temperature), which can be resource-intensive. LLM querying is computationally costly, particularly for large graphs like Neuropathic. Token limits restrict the ability to encode full metadata, particularly in complex graphs like Neuropathic, leading to incomplete or inconsistent responses. Specialized variable names may be misinterpreted, especially at higher temperatures, introducing semantic errors and prediction noise. These factors collectively pose scalability and usability challenges, particularly for resource-constrained settings.

\paragraph{Reproducibility of LLM-based CD.}  
While our method is LLM-agnostic, we observe discrepancies with results from \citet{jiralerspong2024efficient}, highlighting broader reproducibility challenges. LLMs are stochastic and sensitive to prompt design, versioning, query order, and API behavior, factors rarely standardized or reported. Without standardized prompting protocols or access to latent states, consistent benchmarking remains difficult.

\paragraph{Risk of Social Bias in LLMs.}  
LLMs trained on web-scale data may encode social biases, leading to causal inferences that reflect stereotypes rather than ground-truth mechanisms. This risks semantic hallucinations and biased edge assignments, particularly in fairness-critical applications. This risk is addressed through statistical weighting and confidence-based filtering, as described in Section \ref{sec:Method}.

\paragraph{Domain and Metadata Dependence.}  
Our framework depends on interpretable variable metadata to guide LLM queries. It performs best in domains like healthcare or census data. In sparse, technical, or ambiguous domains (e.g., genomics, sensor data), performance degrades. While LLM priors scale efficiently, human-in-the-loop priors could be incorporated for low-interpretability domains. Future work could explore hybrid prompting strategies or domain adaptation techniques to extend semantic CD to less interpretable or evolving data environments.

\section{Conclusion}

We propose a fairness-driven CD framework that integrates LLMs with AL. Across benchmarks including synthetic and semi-synthetic real-world networks, our method outperforms prior baselines in structural accuracy and reduces false positives in fairness-critical path recovery. By combining statistical dependencies with LLM-based semantic priors and prioritizing informative queries, our framework improves both reliability and fairness-awareness in CD.

To support reproducible evaluation, we introduce a semi-synthetic benchmark based on the UCI Adult dataset, embedding structural bias, latent confounding, and noise. This provides a realistic yet controlled testbed with ground-truth graphs for consistent comparison across CD methods.

Our analysis shows that exploration budget (e.g., number of iterations) significantly affects performance, with the relative importance of LLM-derived versus statistical signals shifting with graph size and complexity. While our method improves fairness-path recovery, it does not offer end-to-end guarantees for downstream fairness outcomes. Instead, it uncovers explainable causal pathways, particularly those involving sensitive attributes intended to guide audits, impact assessments, or stakeholder review.

This positions our framework as an improvement in diagnostic tools for fairness evaluations. Use cases include hiring audits, policy design, and regulatory compliance reporting. By recovering interpretable causal paths, our method offers a step toward tools that support fairness-focused exploration, potentially informing future integration into dashboards or algorithmic auditing pipelines. We plan to extend the framework to dynamic graphs and release open-source pipelines for scalable, reproducible fairness audits.

\subsubsection*{Ethical Statement.}
This work advances fairness-aware causal discovery by proposing a framework to identify unfair pathways from sensitive attributes to outcomes. The approach can inform the development of more equitable algorithms and policy interventions. However, causal inferences drawn from observational data have inherent limitations and may be misinterpreted without domain expertise. We caution against over-reliance on LLM judgments in high-stakes settings and stress that our method is intended for diagnostic and research purposes only. Additionally, the use of LLMs incurs environmental costs due to their energy-intensive inference. Future work should explore more efficient, bias-aware models to reduce this impact.

\section*{Acknowledgments}
This study was funded by NSF \#2047296 and OpenAI’s Researcher Access Program.

\bibliography{aaai25}

\appendix
{

\setcounter{secnumdepth}{2}  

\section{Synthetic Data Generation}
\label{app:syn_data}

This section provides more detail on the Adult-based synthetic data general rules, formulas, and process.The graph was constructed based on causal assumptions documented in prior fairness research \cite{kilbertus2017avoiding, nabi2018fair}. Variable relationships were chosen to reflect widely accepted socioeconomic patterns and to ensure transparency and reproducibility, minimizing author-imposed bias.

\subsection{Data Generation Process}
\label{app:data_gen}

We generate synthetic data based on socioeconomic variables from the UCI Adult dataset. Each variable is sampled according to structural equations encoded in a DAG. Variables are simulated in topological order using a JSON-defined structural causal model with dependency-specific functional forms. The outcome variable \texttt{income} is modeled as a non-linear threshold function over upstream predictors (e.g., capital gain, years of education) and sensitive attributes, mimicking realistic socioeconomic determinants. Data are generated using a Python pipeline with sample sizes of 15,000 and 50,000. Each variable is sampled according to structural equations encoded in a DAG. Table \ref{tab:baseline-generation} summarizes the variable types, parent dependencies, and sampling formulas included in the baseline graph generation. 

\begin{table*}[ht] 
\caption{Variable generation rules for the synthetic dataset (baseline).}
\label{tab:baseline-generation} 
\begin{center}
\begin{scriptsize}
\begin{tabular}{|p{2.5cm}|p{2.2cm}|p{3.8cm}|p{5.5cm}|} 
\toprule

\textbf{Variable} & \textbf{Type} & \textbf{Dependencies} & \textbf{Generation Formula} \\
\midrule

\texttt{age} & Numerical & None & {Uniform}(18, 70) \\

\texttt{sex} & Categorical & None & ["Male", "Female"], p = [0.67, 0.33] \\

\texttt{race} & Categorical & None & ["White", "Non white"], p = [0.75, 0.25] \\ 

\texttt{native country} & Categorical & None & 5 categories, p = [0.75, 0.1, 0.05, 0.05, 0.05] \\ 

\texttt{education} & Categorical & sex, race, native country & Conditional rules on demographics \\

\texttt{years of education} & Numerical & education, age & \ensuremath{\mathcal{N}(12 + 2 \cdot \text{education} + 0.05 \cdot \text{age}, 1)} \\ 

\texttt{occupation} & Categorical & education, sex & Rule-based mapping by sex and education \\ 

\texttt{work class} & Categorical & education, race & Rule-based mapping by education and race \\ 

\texttt{marital status} & Categorical & age, sex & Rule-based by age thresholds and sex \\ 

\texttt{relationship} & Categorical & marital status & Rule-based mapping from marital status \\ 

\texttt{capital gain} & Numerical & occupation & Normal based on occupation \\ 

\texttt{capital loss} & Numerical & occupation & Normal based on occupation \\

\texttt{work hours per week} & Numerical & occupation, sex & Normal with group-dependent means \\ 

\texttt{fnlwgt} & Numerical & None & \ensuremath{\mathcal{N}(180000, 20000)} \\ 

\texttt{income} & Categorical & all above & Conditional rule combining capital gain, years of education, work hours per week, sex, race \\
\bottomrule

\end{tabular}
\end{scriptsize}
\end{center}
\end{table*}

Table \ref{tab:noisy-generation} configuration describes the synthetic data generation process with noise and latent confounding. All formulas reflect variable dependencies and data-generation logic used in the noisy setup. Here, $U \sim \mathcal{N}(0, 1)$ denotes the unobserved latent confounder included in select formulas to simulate realistic social or institutional bias.

\begin{table*}[ht] 
\caption{Noisy Synthetic Data Generation Configuration.}
\label{tab:noisy-generation} 
\begin{center}
\begin{scriptsize}
\begin{tabular}{|p{2.2cm}|p{3cm}|p{3cm}|p{5.5cm}|} 
\toprule

\textbf{Variable} & \textbf{Type} & \textbf{Dependencies} & \textbf{Formula / Sampling Rule} \\
\midrule

\texttt{age} & Numerical & None & \ensuremath{\mathcal{U}(18, 70)} \\

\texttt{sex} & Categorical (Male, Female) & None & Probabilities: [0.67, 0.33] \\

\texttt{race} & Categorical (White, Non white) & None & Probabilities: [0.75, 0.25] \\ 

\texttt{native country} &  Categorical (US, Mexico, India, Philippines, Other) & None & Probabilities: [0.75, 0.10, 0.05, 0.05, 0.05] \\ 

\texttt{education} & Categorical (5 levels) & sex, race, native country & \texttt{0 if sex==1 and race==1; 1 if native country \ensuremath{\neq} 0; 2 if sex==0 and race==0; else 1} \\

\texttt{years of education} & Numerical (Discrete) & education, age & \ensuremath{\mathcal{N}(12 + 2 \cdot \texttt{education} + 0.05 \cdot \texttt{age} + 0.8 \cdot U,\ 1)} \\ 

\texttt{occupation} & Categorical (5 classes) & education, sex & \texttt{2 if education < 1; 1 if sex==0 and education \ensuremath{\geq} 2; 0 if education \ensuremath{\geq} 3; else 3} \\ 

\texttt{work class} & Categorical (5 classes) & education, race & \texttt{0 if education > 2; 1 if race==0; 2 if education \ensuremath{\leq} 1; else 4} \\ 

\texttt{marital status} & Categorical (5 classes) & age, sex & \texttt{0 if age < 30; 1 if sex==0 and age \ensuremath{\geq} 30; 2 if sex==1 and age \ensuremath{\geq} 30; else 3} \\ 

\texttt{relationship} & Categorical (4 classes) & marital status & \texttt{0 if marital status==1; 1 if marital status==3; 2 if marital status==0; else 3} \\ 

\texttt{capital gain} & Numerical & occupation & \ensuremath{\mathcal{N}(5000 + 1500 \cdot U,\ 2000)} if \texttt{occupation==1}; else \texttt{1000 + 500 \ensuremath{\cdot} U} or 0  \\ 

\texttt{capital loss} & Numerical & occupation & \ensuremath{\mathcal{N}(1000,\ 500)} if \texttt{occupation==1}; else 300 or 0 \\

\texttt{work hours per week} & Numerical & occupation, sex & \ensuremath{\mathcal{N}(45 + 2 \cdot U,\ 5)} if \texttt{sex==0}; else \texttt{40 + U} or 35 \\ 

\texttt{final weight} & Numerical & None & \ensuremath{\mathcal{N}(180000,\ 20000)} \\ 

\texttt{income} & Categorical (<=50K, >50K) & sex, race, education, years of education, occupation, work class, capital gain, capital loss, work hours per week, relationship, marital status, final weight & \texttt{1 if sex==0 and race==0 and capital gain + 500 \ensuremath{\cdot} U > 4000 and years of education + 0.5 \ensuremath{\cdot} U \ensuremath{\geq} 14 and work hours per week + U > 40; else 0} \\

\bottomrule

\end{tabular}
\end{scriptsize}
\end{center}
\end{table*}

By explicitly modifying only a few formula components (education, gain, hours, income) and keeping the DAG structure constant, we enable controlled experimentation on how hidden confounding affects CD. This setup supports fairness-sensitive benchmarking while maintaining realism and interpretability.

The baseline synthetic network generated is shown in Figure \ref{fig:base_causal_graph}.

\begin{figure}[ht]
    \centering
    \includegraphics[width=\linewidth]{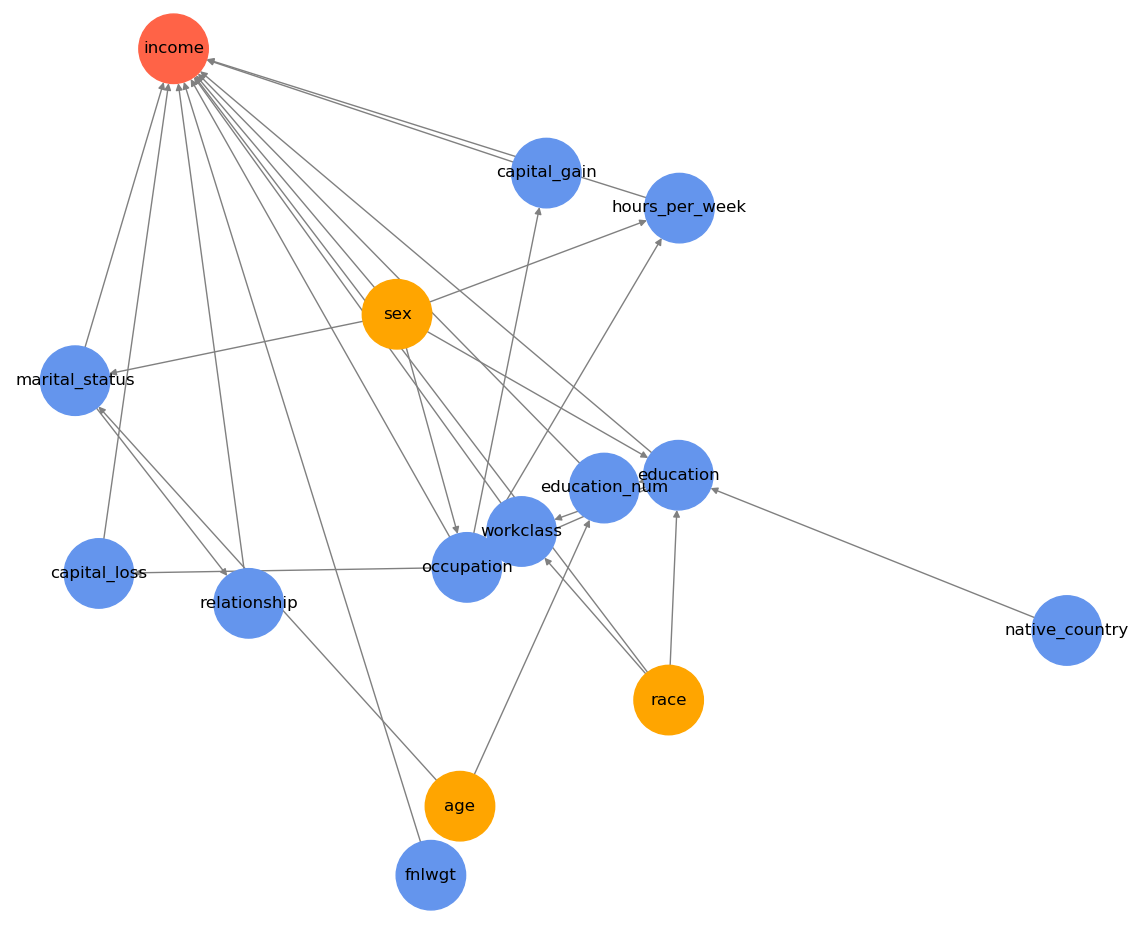}
    \caption{Baseline Causal Graph for Adult-based Synthetic Network. Red and orange nodes represent target and sensitive attributes respectively, and blue nodes represent others.}
    \label{fig:base_causal_graph}
\end{figure}

\subsection{Latent Confounding via Unobserved Variable $U$} 
\label{app:latent_variable}

To reflect unmeasured societal influences (e.g., privilege, institutional access), we introduce a latent confounder $U \sim \mathcal{N}(0, 1)$, that affects intermediate variables such as years of education, capital gain, and hours of work per week, thereby indirectly influencing income. This reflects latent societal advantages that are not captured in observed features.

The variables influenced by $U$ were selected based on evidence from the fairness literature and sociological theory. For example, years of education is shaped by early-life educational resources and family background \cite{nabi2018fair, pearl2009causality}. Similarly, capital gain reflects socioeconomic access to investments, and work hours are constrained by latent factors like flexibility or family obligations \cite{nabi2018fair, pearl2009causality, hardt2016equality, heckman2005inequality}.

To simulate these hidden biases, each equation embeds the latent variable $U$ as an additive linear term, following standard structural causal model (SCM) assumptions under additive noise models (ANMs) \cite{wu2021counterfactual, kilbertus2017avoiding}. This models $U$ as an unobserved common cause influencing multiple downstream variables.

The latent confounder is injected via additive linear terms into structural equations following standard assumptions in structural causal models (SCMs). For the binary outcome income, we use a nonlinear threshold function that combines observable and latent contributions. This setup reflects bias amplification near decision boundaries and is consistent with prior fairness simulation frameworks.

Continuous variables are modeled using linear structural equations. Gaussian additive noise models of the form $X_i = f(\text{Parents}_i) + \beta \cdot U_i + \epsilon_i,\ \epsilon_i \sim \mathcal{N}(0, \sigma^2)$ are used. Here, $X_i$ is an observed variable such as \textit{ years of education} or \textit{capital gain}; $\mu(\cdot)$ represents the deterministic contribution of observed parent variables (e.g., \textit{education}, \textit{age}); $\beta \cdot U_i$ encodes the latent influence of the unobserved variable $U$; and $\epsilon_i$ is a zero-mean residual. For instance, the variable years of education is generated as $\text{Years of Education} \sim \mathcal{N}(12 + 2\,\text{education} + 0.05\,\text{age} + 0.8\,U,\ 1)$.

This formulation enables realistic partial confounding, allowing $U$ to simulate latent social advantage while preserving identifiability.It aligns with the LiNGAM framework \cite{shimizu2006linear}, structural equation modeling \cite{pearl2009causality}, and recent fairness simulation approaches involving latent variables \cite{kiciman2023causal}.

For binary outcomes, such as income, thresholding logic is implemented over both observed and latent inputs using the equation:

\begin{equation}
Y = \mathbb{I}\left[ f(X) + \lambda U > \theta \right]
\end{equation}

where $f(X)$ is a (possibly nonlinear) function of observable predictors, $\lambda$ is the strength of the latent effect, and $\theta$ is a threshold. The indicator function $\mathbb{I}[\cdot]$ outputs 1 when the threshold is exceeded. In our setup, this is instantiated as:

\begin{align*}
\text{Income} = \mathbb{I}\Big[ &\;\text{capital\_gain} + 500 \cdot U > 4000 \\
&\land\; \text{education\_num} + 0.5 \cdot U \geq 14 \\
&\land\; \text{hours\_per\_week} + U > 40 \Big]
\end{align*}

This design captures bias amplification, where latent privilege encoded by $U$ influences outcomes near decision boundaries. Such threshold-based models are commonly used in fairness simulations to mimic opaque real-world decisions \cite{nabi2018fair, kilbertus2017avoiding, wu2021counterfactual, kiciman2023causal}.

\subsection{Structured Noise Augmentation}

To test robustness under realistic conditions, we augment the clean synthetic data with structured noise and confounding, emulating sources of measurement error and label corruption common in real-world datasets.

\paragraph{Additive Noise for Continuous Variables.}

Gaussian noise is added to all continuous variables post-generation to simulate measurement variability.

For each continuous variable X, noise is sampled as:

\begin{equation}
    X_i^{\text{noisy}} = X_i + \epsilon_i, \quad \epsilon_i \sim \mathcal{N}(0, \sigma_X^2), \quad \sigma_X = \alpha \cdot \mathrm{std}(X)
\end{equation}

where $X_i$ is the original value for sample $i$, $\epsilon_i$ is a random error drawn from a normal distribution with mean 0 and standard deviation $\sigma_X$, $\sigma_X=\alpha \cdot \mathrm{std}(X)$, where $\alpha \in [0, 1]$ is a tunable noise scaling factor set to $\alpha = 0.1$. This maintains proportionality between noise magnitude and feature variability, consistent with additive noise models used in ICA-LiNGAM and structural equation modeling (SEM) frameworks \cite{shimizu2006linear, bollen1989structural}.

\subsubsection{Symmetric Label Noise in Categorical Variables}

To emulate mislabeling, we introduce symmetric label noise into categorical variables. For each categorical variable, a proportion  $\gamma$ of values (5\%) are randomly replaced with an alternative category, drawn uniformly from the remaining labels. For each categorical variable $Y$ with $K$ classes (e.g., "Male", "Female"), uniform symmetric noise is introduced using the following equation.

Let $\gamma \in [0,1]$ be the label noise probability. Then for each sample $i$:

\begin{equation}
P(\tilde{Y}_i = y') = 
\begin{cases}
1 - \gamma & \text{if } y' = Y_i \\
\frac{\gamma}{K - 1} & \text{if } y' \ne Y_i
\end{cases}
\end{equation}

Where $Y_i$ is the original category label,  
$\tilde{Y}_i$ is the observed (possibly noisy) label, $K$ is the number of possible categories for $Y$, and $y' \in \mathcal{Y}$ are all possible class values.

This means with probability $\gamma$, the label is flipped uniformly to one of the $K - 1$ incorrect categories. And with probability $1 - \gamma$, the label remains correct. This uniform corruption model known as symmetric label noise, is widely used in label-noise robust learning \cite{natarajan2013learning, patrini2017making}.

\subsubsection{Controlled Replication Across Seeds}

To evaluate generalization and fairness robustness, we generate four datasets using different random seeds (42, 101, 222, 333), each with the same underlying graph but distinct noise realizations. Noise is added post-graph generation, preserving causal structure while introducing rea)listic variability. This mirrors standard practices in fairness benchmarking and robustness testing for CD.

\section{Experimental Details} \label{app:experiments_details}

In this section, we detail the specific hyperparameters utilized for each set of experiments.

\paragraph{Synthetic Adult-based Network.} We apply six baseline methods (PC, GES, NOTEARS, DAGMA, LLM Pairwise, and LLM BFS) to each dataset instance and averaged performance metrics across seeds. For NOTEARS and DAGMA, we conduct a sweep over multiple regularization values ($\lambda \in \{0.001, 0.005, 0.01, 0.05, 0.1\}$) and selected the best performing configuration for reporting: $\lambda=0.01$ for NOTEARS and $\lambda=0.05$ for DAGMA. 

We perform 200 trials per dataset, optimizing for F1 score via Bayesian search. The tuned hyperparameter space includes: query prioritization weights selected from $[(0.3, 0.3, 0.4), (0.25, 0.25, 0.5), (0.4, 0.3, 0.3), (0.3, 0.4, 0.3)]$, score threshold values between 0.25 and 0.35, temperature values between 0.25 and 0.6, and a maximum iteration range of 40 to 100. We report the average across seeds for all performance metrics used in evaluation.

Fairness pathway analysis compares recovered and true graphs to assess whether each method correctly identifies bias mechanisms. CD outputs include (1) Causal Graphs: Represented by adjacency matrices capturing the graph structure, and (2) Causal Dictionaries: Capturing identified causal relationships. We focus on paths from sensitive attributes (sex, race, age) to the outcome variable (income) for fairness evaluation.

An example of a full prompt for this experiment is listed in Listing \ref{lst:prompt-synadult}.

\begin{listing}[H]
\caption{LLM Query Example - Syn-Adult Dataset}
\label{lst:prompt-synadult}
\begin{lstlisting}
System: You are an expert in socioeconomic and demographic research.
User: You are a helpful assistant to an expert in socioeconomics, with a focus on income prediction and labor statistics.

Metadata:
Variable A: education: highest level of education achieved (e.g., Bachelors, Masters)
Variable B: income: income level (whether the individual's income exceeds \$50K/year) ...

Query:
Does education cause income?
Respond with <Answer>Yes</Answer> or <Answer>No</Answer>.
\end{lstlisting}
\end{listing}

\paragraph{Child Network.} We perform 1000 trials across all chunks for the proposed method. Evaluation is based on the F1 score with respect to the ground-truth DAG. Hyperparameter ranges are: score threshold $[0.01, 0.2]$, temperature $[0.05, 0.7]$, and max iterations $[10, 50]$. 

An example of a full prompt for this experiment is listed in Listing \ref{lst:prompt-child}.

\begin{listing}[H]
\caption{LLM Query Example - CHILD Dataset}
\label{lst:prompt-child}
\begin{lstlisting}
System: You are an expert on children's diseases.
User: You are a helpful assistant to a children's disease expert.

Metadata:
Variable A: BirthAsphyxia: lack of oxygen to the blood during the infant's birth
Variable B: Disease: infant methemoglobinemia

Query:
Does BirthAsphyxia cause Disease?
Respond with <Answer>Yes</Answer> or <Answer>No</Answer>.
\end{lstlisting}
\end{listing}

\paragraph{Neuropathic Pain Network.} We run 200 extended trials for the proposed method to explore effects on long-range dependencies. Parameter ranges include score thresholds $[0.1, 0.5]$, temperature $[0.1, 0.3]$, and max iterations $[200, 500]$.

An example of a full prompt for this experiment is listed in Listing \ref{lst:prompt-neuropathic}.

\begin{listing}[H]
\caption{LLM Query Example - Neuropathic Dataset}
\label{lst:prompt-neuropathic}
\begin{lstlisting}
System: You are an expert on neuropathic pain diagnosis.
User: You are a helpful assistant to a neuropathic pain diagnosis expert.

Metadata:
Variable A: L C6 Radikulopati: left C6 radiculopathy (nerve pain originating at cervical vertebra C6)
Variable B: L Handbesvar: left hand discomfort

Query:
Does L C6 Radikulopati cause L Handbesvar?
Respond with <Answer>Yes</Answer> or <Answer>No</Answer>.
\end{lstlisting}
\end{listing}

\section{Parameter Impact Analysis on Proposed Method For Child and Neuropathic Networks}
\label{app:param_analysis}

To evaluate the influence of various parameters on the performance of the proposed method, a detailed analysis of the hyperparameters is conducted using visualizations such as ridge plots, correlation heatmaps, and feature importance analysis.

\paragraph{Impact of Weights on Performance.} Figure \ref{fig:ridge_plots} shows ridge plots illustrating the impact of the 3-weight combination on the F1 score for experiments on the Child network. Each plot visualizes F1 score distributions across different weight bins. The width indicates variation in F1 scores; a wider ridge means more variation, while a narrower ridge suggests consistent performance. The height of the peak shows the most frequently occurring F1 scores, with a taller peak indicating a common F1 score for that weight bin. Colors distinguish ridges for each bin without indicating performance or significance. The figure shows that statistical information and query history are significant contributors to performance, while LLM confidence has a stable, less pronounced effect.

\begin{itemize}
    \item Statistical Information Weight: Lower bins (0.0-0.1) show narrower ridges and lower F1 scores. As the weight increases (0.3-0.6), distributions broaden, and peaks shift toward higher F1 scores, suggesting optimal performance in this range. Beyond 0.6, performance flattens with diminishing returns.

\item LLM Confidence Weight: Distributions remain stable across all bins, with minor improvements in midrange bins (0.3-0.5). This weight has a minor impact on the F1 score and serves as a supporting factor rather than a decisive one.

\item Query History Weight: Lower bins (0.0-0.2) show lower F1 scores. As the weight increases (0.4-0.6), distributions widen, and peaks shift toward higher F1 scores. Beyond 0.6, improvement plateaus, with optimal performance around 0.4-0.6.
\end{itemize}

These results emphasize the importance of balancing weights, as increasing one decreases the others due to the constraint that all weights must sum to 1.

\begin{figure}[ht]
\centering
\includegraphics[width=0.9\columnwidth]{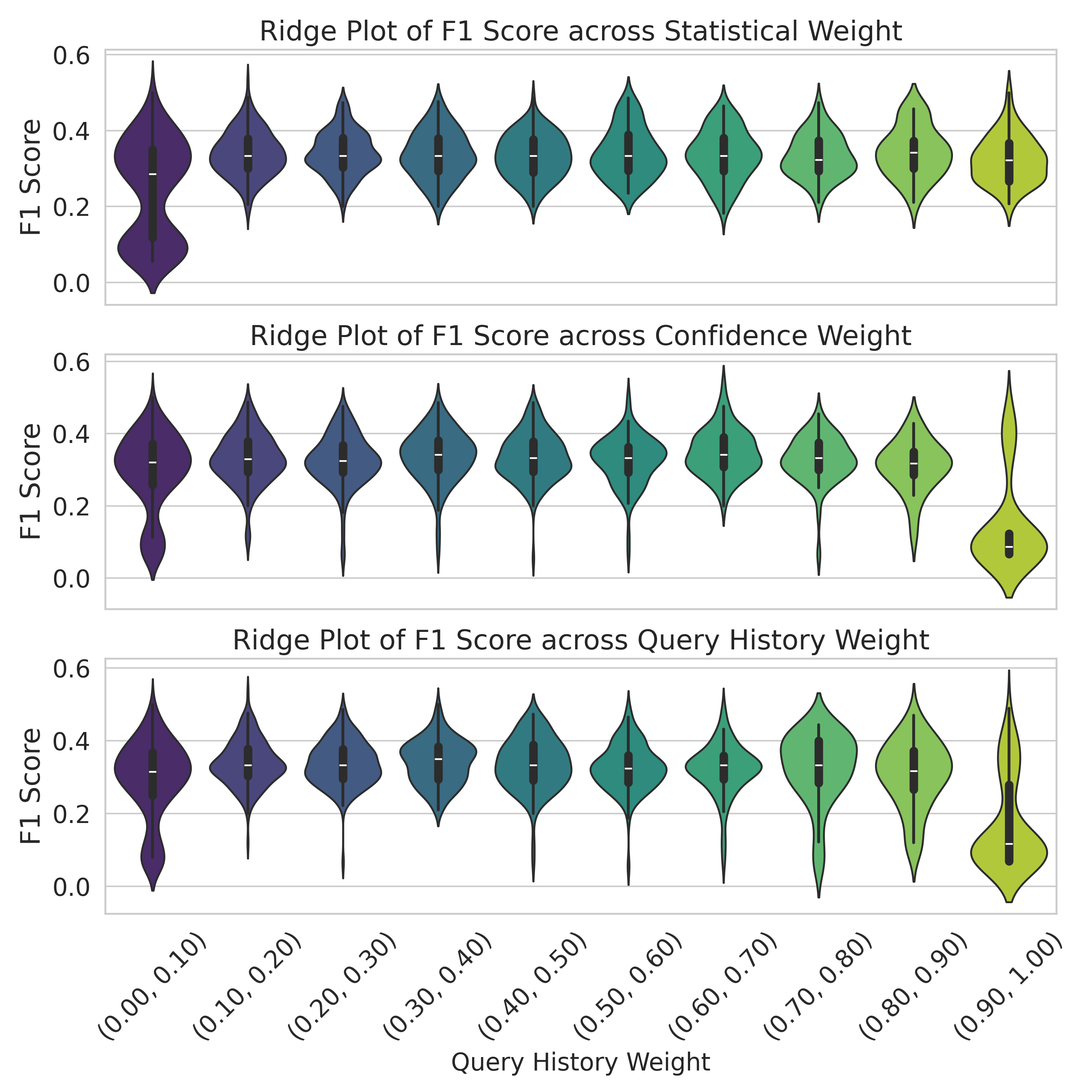} 
\caption{Ridge Plots for Weights of Statistical Information, LLM Confidence, and Query History against F1-Score}
\label{fig:ridge_plots}
\end{figure}

\begin{figure}[ht]
    \centering
    \includegraphics[width=\linewidth]{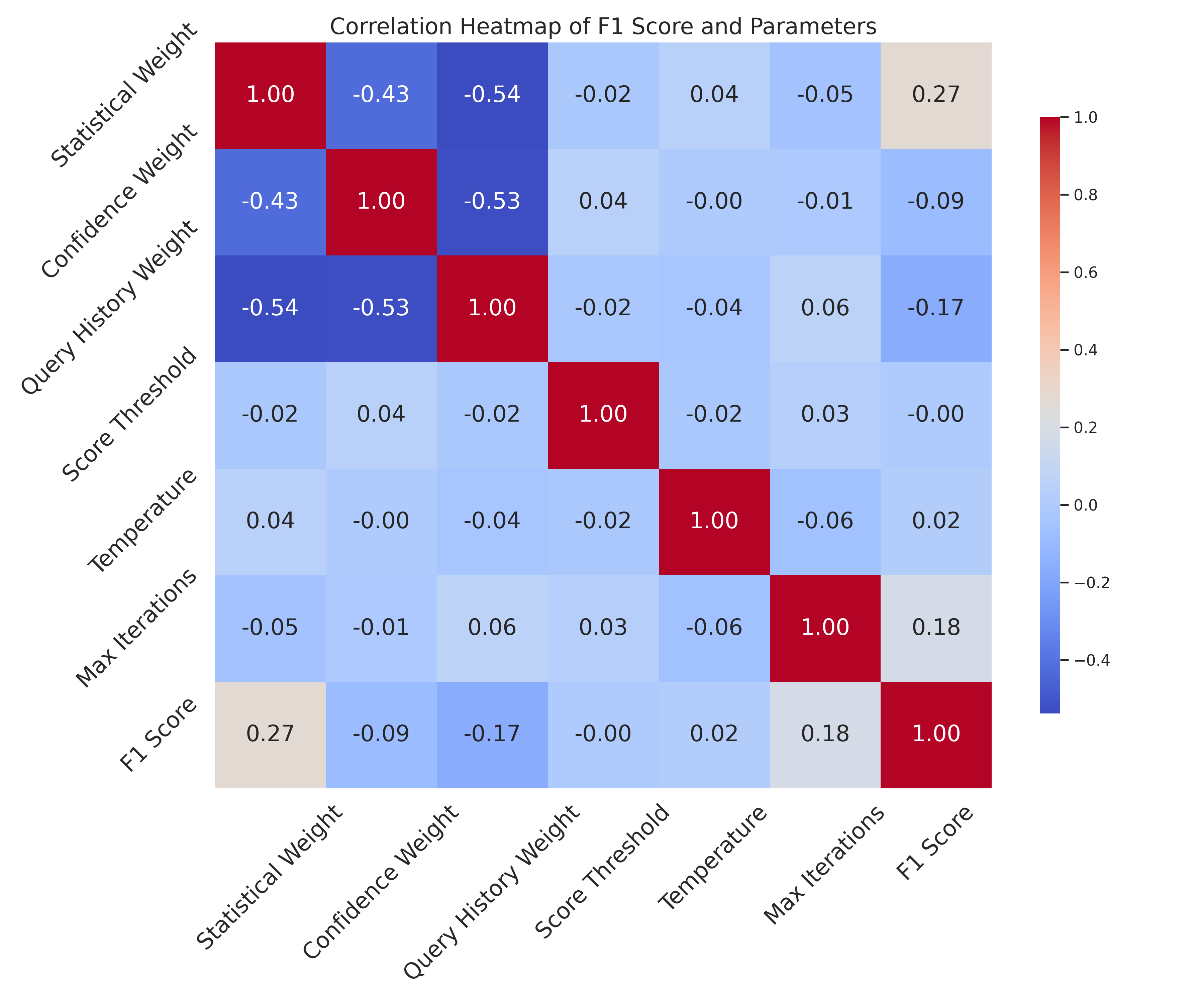}
    \caption{Correlation Heatmap of Parameters and F1 Score - Neuropathic Network}
    \label{fig:corr-heatmap-neuro}
\end{figure}

\paragraph{Correlation Analysis.} Figure \ref{fig:corr-heatmap-neuro} illustrates the relationships between model parameters and the F1 score, revealing key performance factors for experiments from the neuropathic dataset. The weight of statistical information has the strongest positive correlation with the F1 score (+0.27), indicating that increasing its value generally improves performance. The weight of the query history shows a weaker negative correlation (-0.17), reflecting diminishing returns beyond its optimal range (0.4-0.6). The weight of the LLM confidence score has a minimal impact (-0.09 correlation), therefore, its optimization is less critical. Negative correlations between weights (e.g., -0.54 between statistical information and query history) highlight trade-offs due to their sum equaling 1. The score threshold shows negligible influence on the F1 score and can be deprioritized. Temperature and the number of maximum iterations have weak positive correlations (+0.02 and +0.18, respectively), so small adjustments have limited impact.

These findings underscore the importance of optimizing statistical information and query history weights while treating other parameters as secondary contributors to model performance.

\section{Fairness and Path Analyses Results}
\label{app:fair_path_results}

This table shows the average fairness and path scores for the baseline synthetic graph and the graphs generated by the CD methods.

\begin{table*}[ht]
\centering
\caption{Fairness and Causal Pathway Analysis across Methods (Averaged across 4 seeds)}
\label{tab:fairness_path_analysis}
\begin{scriptsize}
\begin{tabular}{l|ccc|c|ccc|c}
\toprule
\textbf{Method} & \textbf{Direct} & \textbf{Indirect} & \textbf{Spurious} & \textbf{Fairness Path} & \textbf{Direct Effect} & \textbf{Indirect Effect} & \textbf{Total Effect} & \textbf{Normalized C\_Bias} \\
 & \textbf{Paths} & \textbf{Paths} & \textbf{Paths} & \textbf{Contribution} &  &  &  & \\
\midrule
\textbf{Baseline (GT)} & 2 & 25 & 25 & 0.52 & 0.623 & 4.272 & 4.894 & 28.465 \\
\textbf{Proposed Method} & 2 & 3 $\pm$ 1 & 5 $\pm$ 1 & 0.467 $\pm$ 0.021 & 0.519 & 0.247 $\pm$ 0.003 & 0.766 $\pm$ 0.002 & 4.602 $\pm$ 0.016 \\
\textbf{LLM BFS} & 3 & 44 $\pm$ 18 & 121 $\pm$ 10 & 0.273 $\pm$ 0.061 & 0.519 & 2.930 $\pm$ 0.47 & 3.447 $\pm$ 0.46 & 20.699 $\pm$ 2.58 \\
\textbf{LLM Pairwise} & 3 & 222 $\pm$ 99 & 281 $\pm$ 114 & 0.443 $\pm$ 0.05 & 0.519 & 6.136 $\pm$ 2.76 & 6.655 $\pm$ 2.66 & 39.992 $\pm$ 15.4 \\
\textbf{NOTEARS (0.01)} & 0 & 0 & 5 $\pm$ 6 & 0.00 & 0.00 & 0.00 & 0.00 & 0.00 \\
\textbf{DAGMA (0.05)} & 0 & 0 & 31 $\pm$ 2 & 0.00 & 0.00 & 0.00 & 0.00 & 0.00 \\
\textbf{GES} & 0 & 45 $\pm$ 10 & 638 $\pm$ 100 & 0.068 $\pm$ 0.03 & $\sim$0 & 2.892 $\pm$ 0.32 & 6.293 $\pm$ 1.81 & 13.975 $\pm$ 7.35 \\
\textbf{PC} & 0 & 1 $\pm$ 1 & 51 $\pm$ 23 & 0.02 $\pm$ 0.01 & 0.00 & 0.221 $\pm$ 0.12 & 0.221 $\pm$ 0.12 & 1.326 $\pm$ 1.05 \\
\bottomrule
\end{tabular}
\end{scriptsize}
\end{table*}

The graphs generated using the proposed method, PC, GES, DAGMA, Notears, LLM Pairwise, and LLM BFS methods are shown in Figures \ref{fig:method_causal_graph} to \ref{fig:bfs_causal_graph}.

\begin{figure}[ht]
    \centering
    \includegraphics[width=\linewidth]{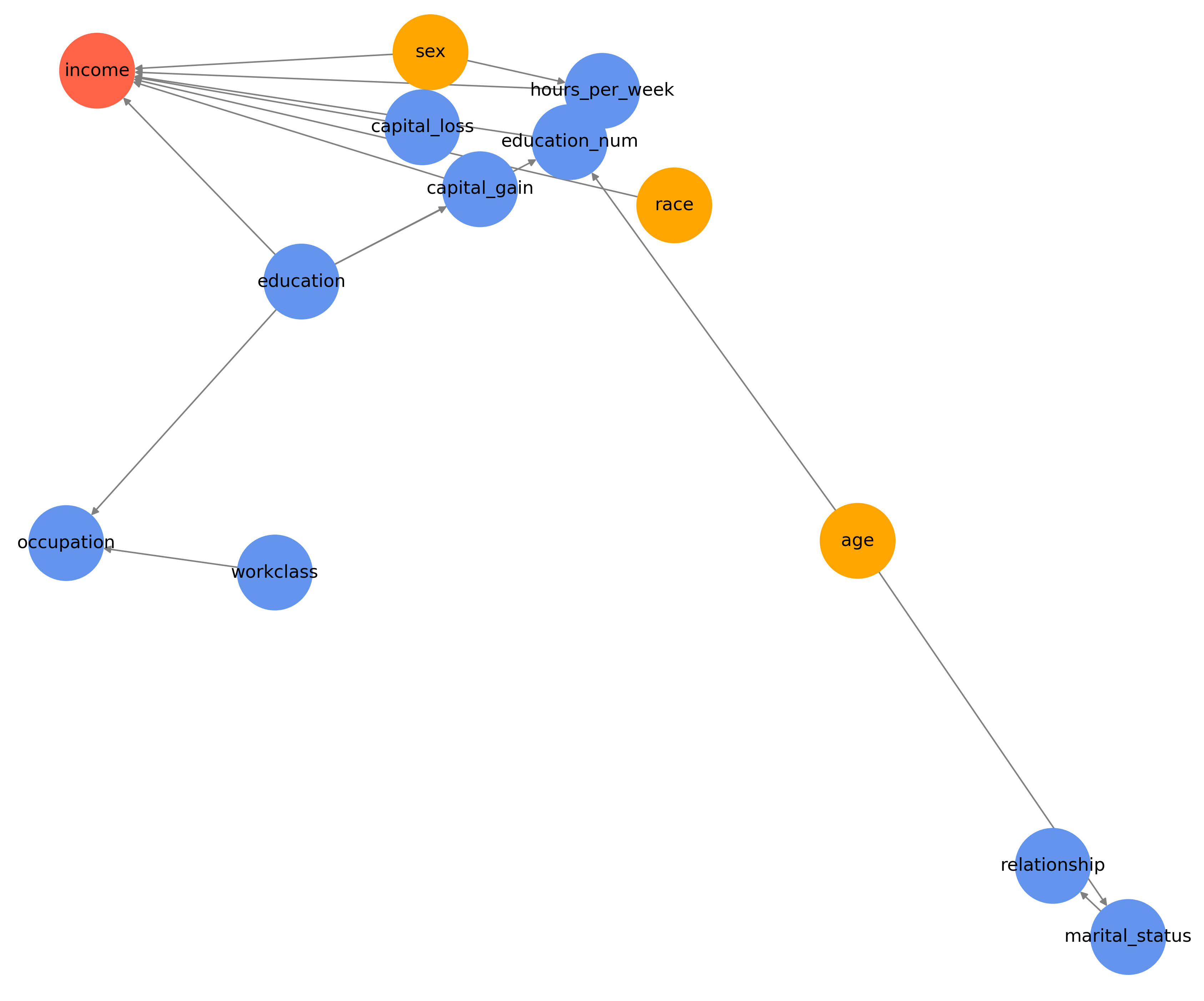}
    \caption{Causal Graph for Adult-based Synthetic Network Generated by Proposed Method. Red and orange nodes represent target and sensitive attributes respectively, and blue nodes represent others.}
    \label{fig:method_causal_graph}
\end{figure}

\begin{figure}[ht]
    \centering
    \includegraphics[width=\linewidth]{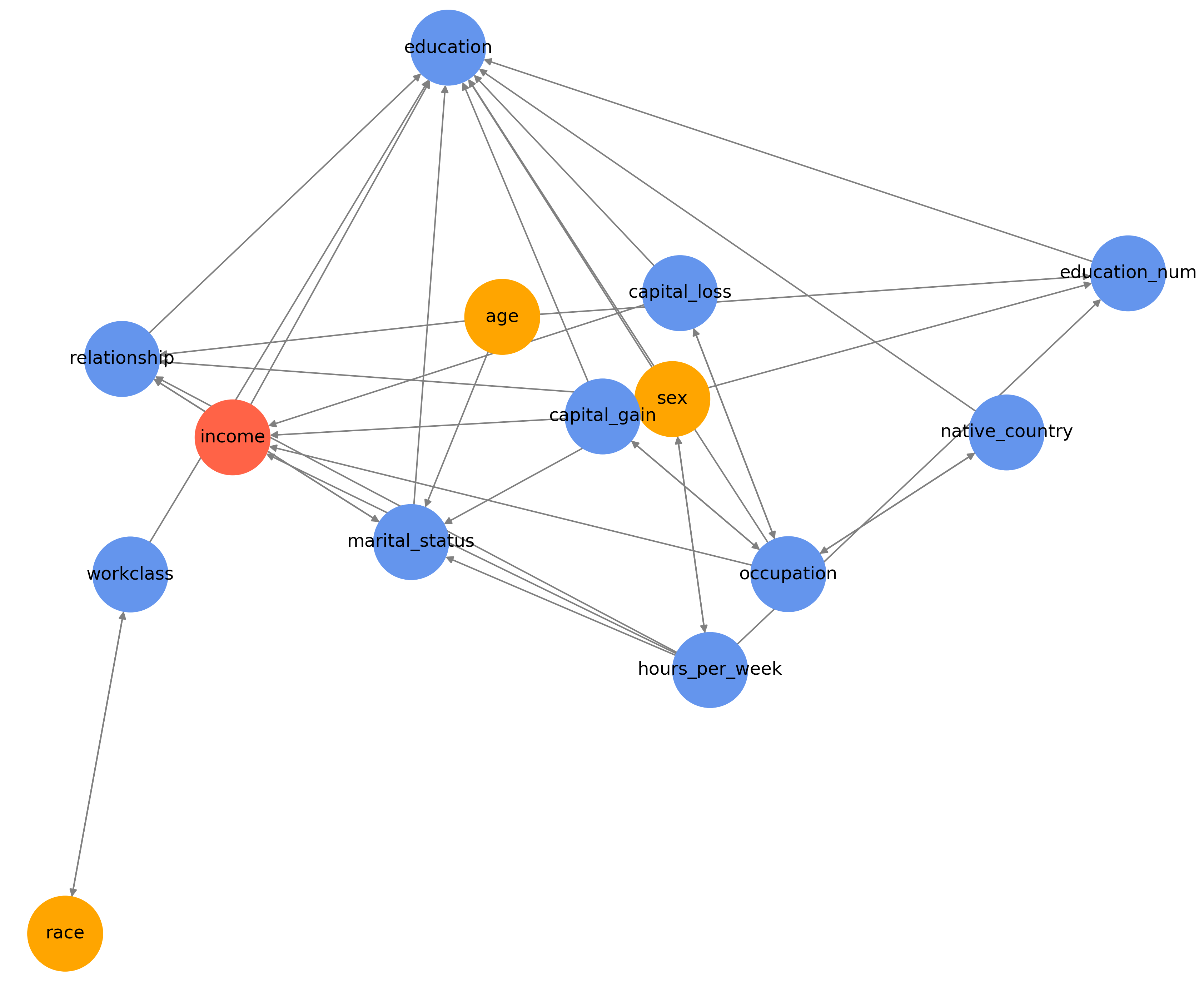}
    \caption{Causal Graph for Adult-based Synthetic Network Generated by PC Method. Red and orange nodes represent target and sensitive attributes respectively, and blue nodes represent others.}
    \label{fig:pc_causal_graph}
\end{figure}

\begin{figure}[ht]
    \centering
    \includegraphics[width=\linewidth]{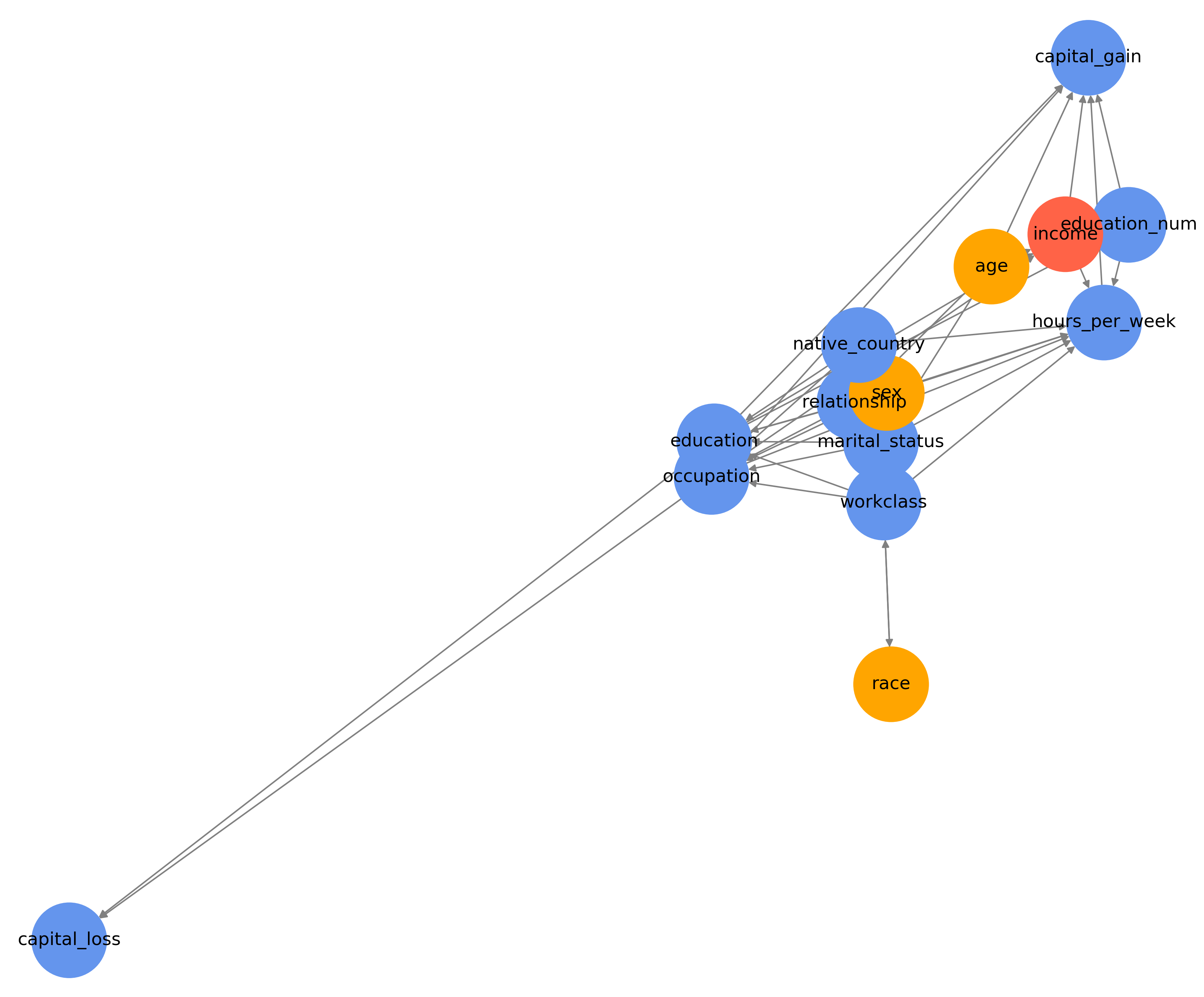}
    \caption{Causal Graph for Adult-based Synthetic Network Generated by GES Method. Red and orange nodes represent target and sensitive attributes respectively, and blue nodes represent others.}
    \label{fig:ges_causal_graph}
\end{figure}

\begin{figure}[ht]
    \centering
    \includegraphics[width=\linewidth]{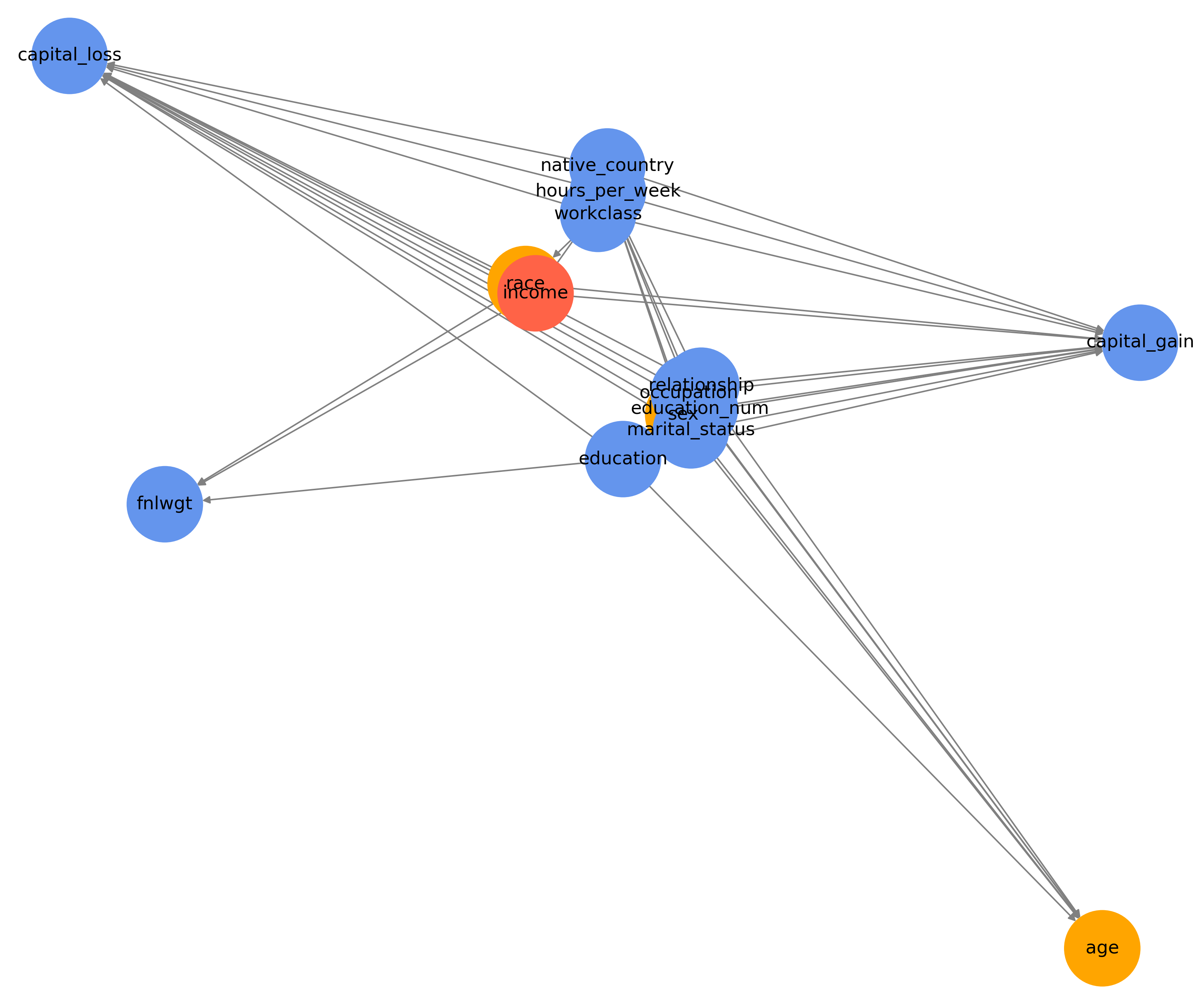}
    \caption{Causal Graph for Adult-based Synthetic Network Generated by Dagma Method. Red and orange nodes represent target and sensitive attributes respectively, and blue nodes represent others.}
    \label{fig:dagma_causal_graph}
\end{figure}

\begin{figure}[ht]
    \centering
    \includegraphics[width=\linewidth]{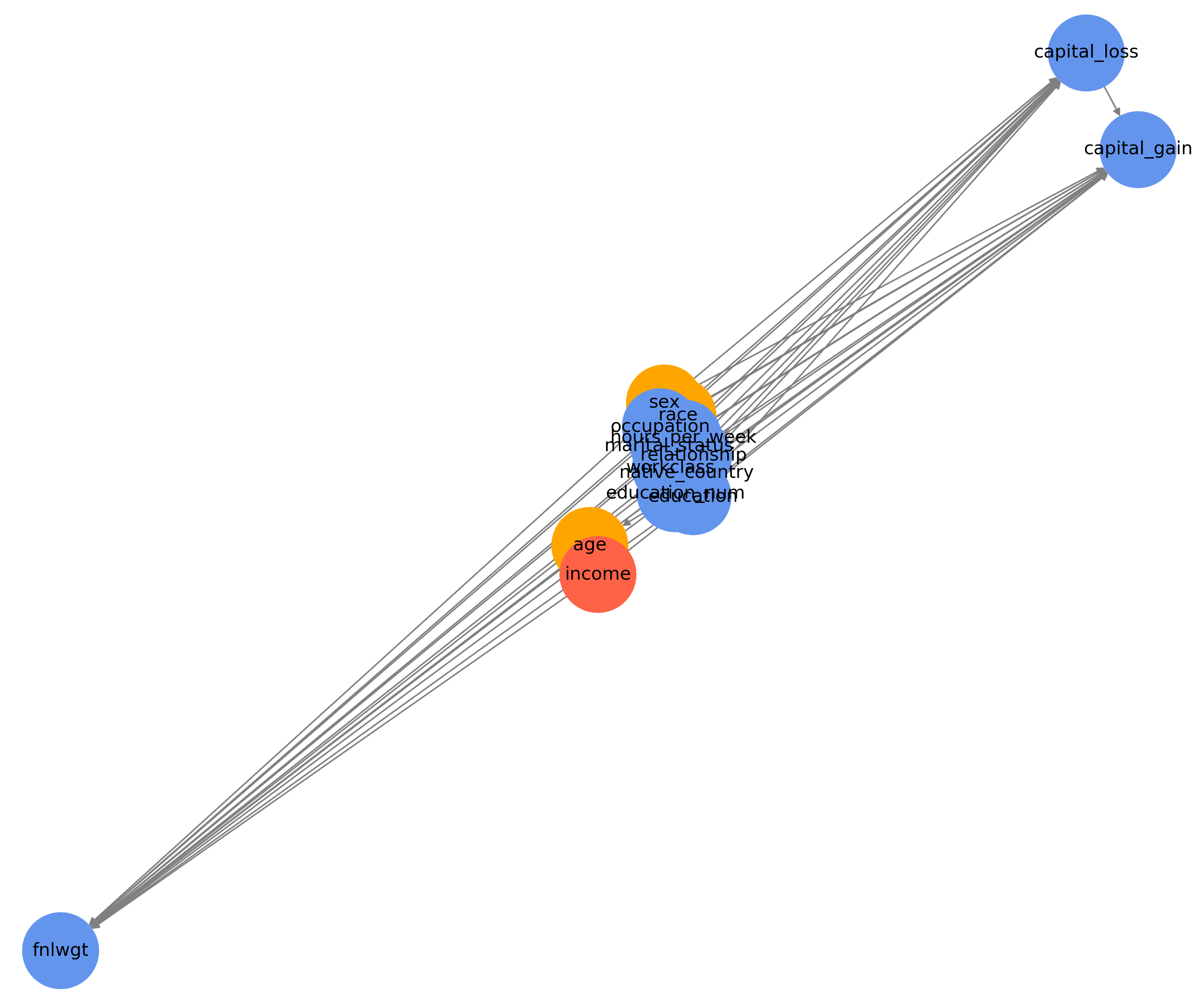}
    \caption{Causal Graph for Adult-based Synthetic Network Generated by Notears Method. Red and orange nodes represent target and sensitive attributes respectively, blue nodes represent others.}
    \label{fig:notears_causal_graph}
\end{figure}

\begin{figure}[ht]
    \centering
    \includegraphics[width=\linewidth]{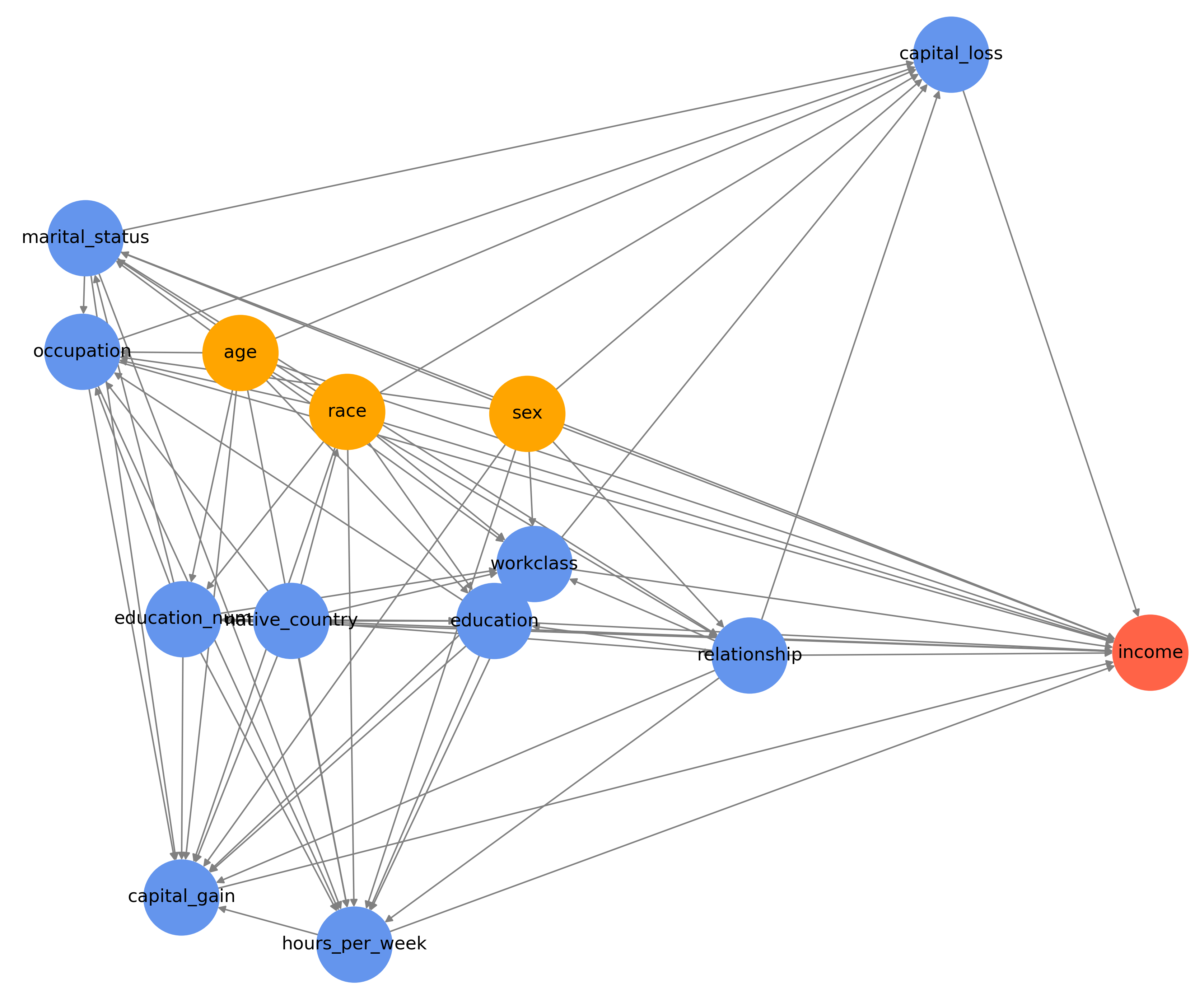}
    \caption{Causal Graph for Adult-based Synthetic Network Generated by LLM Pairwise Method. Red and orange nodes represent target and sensitive attributes respectively, and blue nodes represent others.}
    \label{fig:pairwise_causal_graph}
\end{figure}

\begin{figure}[t]
    \centering
    \includegraphics[width=\linewidth]{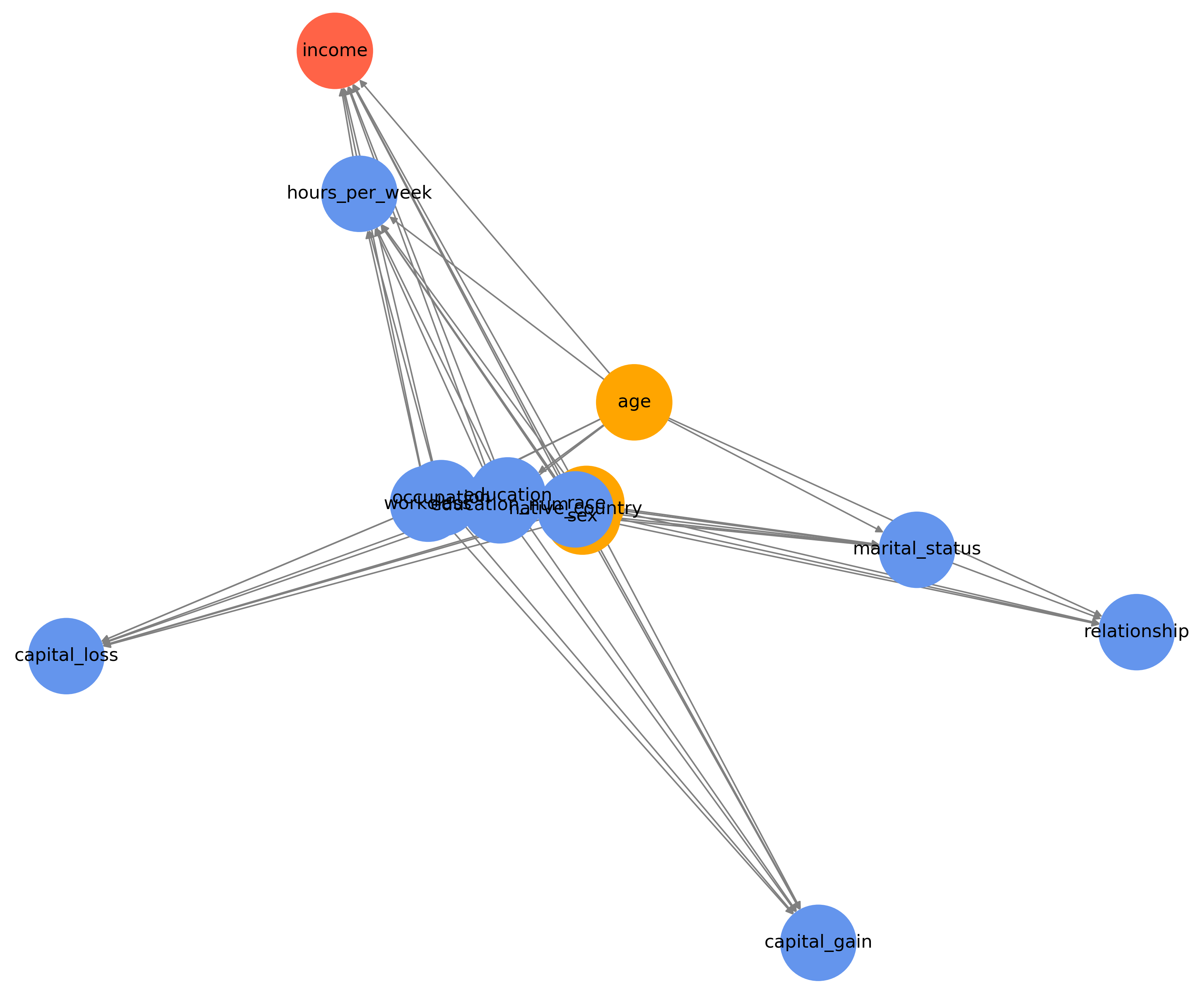}
    \caption{Causal Graph for Adult-based Synthetic Network Generated by LLM BFS Method. Red and orange nodes represent target and sensitive attributes respectively, and blue nodes represent others.}
    \label{fig:bfs_causal_graph}
\end{figure}

\section{Complexity Analyses} \label{app:complexiy_analyses}

The original BFS method \cite{jiralerspong2024efficient} exhaustively explores variable relationships without prioritization, resulting in high computational complexity. It queries independent nodes with complexity \( \mathcal{O}(n) \), where \( n \) is the number of variables. For each variable \( x \), the LLM is queried about all potential children \( y \), leading to a total query complexity of \( \mathcal{O}(n^2) \). Cycle prevention has complexity \( \mathcal{O}(n + e) \), where \( e \) is the number of edges, resulting in a worst-case overall complexity of \( \mathcal{O}(n^3) \).

The proposed method introduces AL and a Dynamic Scoring Mechanism to reduce the number of queries. Querying independent nodes remains \( \mathcal{O}(n) \). Pre-computing scores for all pairs \( (x, y) \) has complexity \( \mathcal{O}(n^2) \). Dynamic scoring during each iteration involves updating scores, with complexity \( \mathcal{O}(k \cdot (n^2 - q)) \), where \( q \) is the number of queried pairs and \( k \) is the maximum number of iterations. This approach prioritizes high-scoring pairs, reducing \( q \) compared to the original BFS method. Cycle prevention complexity is \( \mathcal{O}(q' \cdot (n + e)) \), where \( q' \ll n^2 \). The overall complexity is $\mathcal{O}(n^2 + k \cdot q' + q' \cdot (n + e))$. This significantly improves efficiency over the original BFS approach. Table \ref{tb:complexities} summarizes these improvements.

\textbf{Definition of \( q \) and \( q' \).} We define:
\begin{itemize}
    \item \( q \): The number of variable pairs actually queried by the LLM.
    \item \( q' = n^2 - q \): The number of unqueried or skipped variable pairs due to low informativeness.
\end{itemize}
The scoring function is updated only for a subset of \( q' \), typically involving nodes affected by newly added edges, thus avoiding global score recomputation.

\begin{table*}[h]
\caption{Computational Complexities of Proposed Method Vs. BFS Method}
\label{tb:complexities}
\begin{center}
\begin{scriptsize}
\begin{tabular}{|c|c|c|c|}
\hline
\textbf{Step} & \textbf{BFS method \cite{jiralerspong2024efficient}} & \textbf{Proposed Method} & \textbf{Improvement} \\
\hline
Independent Nodes & $O(n)$ & $O(n)$ & No improvement \\
\hline
Querying Relationships & $O(n^2)$ & $O(k \cdot q')$ & Reduces queries through prioritization. \\
\hline
Dynamic Scoring & Not applicable & $O(n^2 + k \cdot q')$ & Prioritizes high-impact queries. \\
\hline
Cycle Detection & $O(n^2 \cdot (n + e))$ & $O(q' \cdot (n + e))$ & Fewer edges evaluated due to early stopping. \\
\hline
Overall Complexity & $O(n^3)$ & $O(n^2 + k \cdot q')$ & Significant reduction in queries. \\

\hline
\end{tabular}
\end{scriptsize}
\end{center}
\end{table*}

}

\end{document}